\documentclass[11pt]{article}

\usepackage[preprint]{acl}

\usepackage{times}
\usepackage{latexsym}

\usepackage[T1]{fontenc}

\usepackage[utf8]{inputenc}

\usepackage{microtype}

\usepackage{inconsolata}

\usepackage{graphicx}

\usepackage{multicol}
\usepackage{multirow}
\usepackage{amsmath}
\usepackage{amsfonts}
\usepackage{booktabs}
\usepackage{subcaption}
\usepackage{enumitem}
\usepackage{url}
\usepackage{xurl}
\usepackage{float}

\usepackage{tcolorbox}

\usepackage[normalem]{ulem}

\newcommand{\dotsocr}{\texttt{dots.ocr}}
\newcommand{\hunyuan}{\texttt{HunyuanOCR}}
\newcommand{\paddle}{\texttt{PaddleOCR-VL}}

\newtcbox{\upbox}{%
  on line,
  boxsep=1pt,
  left=2pt,right=2pt,
  top=0pt,bottom=0pt,
  colback=green!8,
  colframe=green!8,
  coltext=green!45!black,
  arc=3pt,
  boxrule=0pt
}

\newtcbox{\downbox}{%
  on line,
  boxsep=1pt,
  left=2pt,right=2pt,
  top=0pt,bottom=0pt,
  colback=blue!8,
  colframe=blue!8,
  coltext=blue!45!black,
  arc=3pt,
  boxrule=0pt
}

\newcommand{\upcell}[2]{%
  #1\,\upbox{\tiny$\uparrow$\makebox[2.2em][r]{#2}}%
}

\newcommand{\downcell}[2]{%
  #1\,\downbox{\tiny$\downarrow$\makebox[2.2em][r]{#2}}%
}

\title{Mobilizing the Archive: Bitext Extraction at Scale from Archival Documents for Extremely Low-resource Languages}

\author{Chutong Meng \\
  George Mason University \\
  \texttt{cmeng2@gmu.edu} \\\And
  Alec J. Bateman \\
  Barron Associates \\
  \texttt{bateman@barron-associates.com} \\\And
  Antonios Anastasopoulos \\
  George Mason University \\
  \texttt{antonis@gmu.edu} \\
  }

\begin{document}
\maketitle

\begin{figure*}[t]
    \centering
    \includegraphics[width=0.7\linewidth]{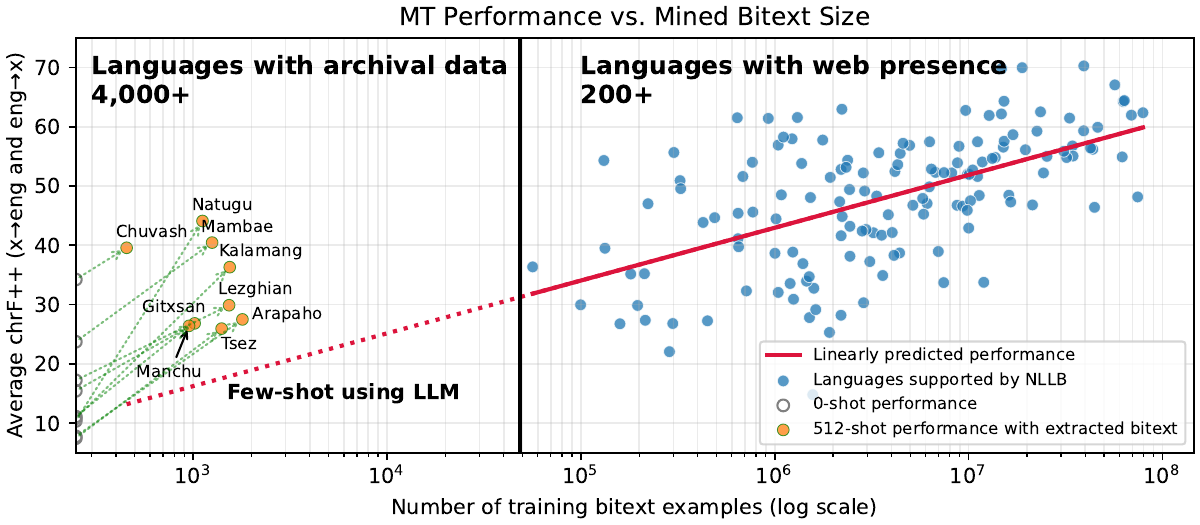}
    \vspace{-1em}
    \caption{We illustrate the rough relationship between mined bitext size and MT performance, using \texttt{nllb-200-3.3B} as an example.
    Archival linguistic resources could extend bitext coverage from around 200 languages to thousands more, while substantially improving few-shot translation quality toward NLLB-level performance.
    }
    \label{fig:motivation}
\vspace{-1.5em}
\end{figure*}

\begin{abstract}
While mining parallel sentences from web pages has proven effective for low-resource machine translation, the majority of the world's languages have little to no web presence.
What they \textit{do} have, though, are resources that are the result of centuries of efforts by linguists, such as grammars, dictionaries, and other documentation materials.
We propose extracting translation examples from such linguistic resources, which offer parallel data for languages under-represented online and could potentially scale to thousands more languages.
We develop a pipeline to achieve this, leveraging an Optical Character Recognition (OCR) model as well as a vision language model (VLM), to ensure identified content to be aligned with visual layout information.
We demonstrate the utility of the extracted bitexts for bootstrapping translation systems across 20 languages, both by training machine translation models and by applying in-context learning for large language models.
For in-context learning, we observe improvements of +15.78 and +6.94 chrF2++ over zero-shot translation for into low- and high-resource languages, respectively.\footnote{Our code and extracted data are publicly available at \url{https://github.com/mct10/mobilizing_the_archive}.}
\end{abstract}

\section{Introduction}

Modern machine translation~(MT) models are typically trained on tens to hundreds of millions of \textit{parallel sentences} (also termed as \textit{bitexts}).
These data are obtained by aligning parallel documents~\cite{koehn-2005-europarl} or mining from large-scale monolingual web data~\cite{schwenk-etal-2021-ccmatrix,nllbteam2022languageleftbehindscaling}.
However, most of the world's languages lack sufficient web data and, consequently, parallel sentences~\cite{joshi-etal-2020-state}, making data scarcity a challenge for underrepresented languages.

Large language models~(LLMs) offer a potential alternative through in-context learning, whereby models can adapt to new tasks from a small number of examples provided in context~\cite{gpt3,wei2022finetuned}.
Recent work has used various linguistic resources, such as grammar book excerpts, dictionaries, morphological analyzers, and interlinear glossed text~(IGT), to instruct LLMs to translate \textit{unseen} low-resource languages~\cite{tanzer2024a, zhang-etal-2024-hire,ramos-etal-2025-grammamt}.
Among these resources, bitext examples appear to provide the largest gains~\cite{aycock2025can}, motivating our focus on extracting parallel examples.

Linguistic resource collections can also provide substantially broader language coverage than web-based bitext mining.
For example, the DReaM corpus~\cite{virk-etal-2020-dream} contains grammar books, language overviews, and dictionaries covering more than 4,000 languages, whereas large-scale bitext mining typically supports only around 200 languages reliably~\cite{nllbteam2022languageleftbehindscaling}.
Although linguistic resources contain fewer parallel sentences, often only a few thousand, we show they can still lead to substantial improvements when used as in-context examples for MT (cf., Fig.~\ref{fig:motivation}).

Extracting structured data from heterogeneous linguistic resources at scale, however, is challenging.
Prior work either relies on resources with predefined machine-readable structure~\cite{hus-anastasopoulos-2024-back, tanzer2024a}, or requires human effort to identify and extract relevant information based on textual and visual formatting cues~\cite{merx-etal-2024-low, aycock2025can}.
Such approaches are difficult to generalize across previously unseen resources with diverse layouts.

\begin{figure}[t]
    \centering
    \includegraphics[width=0.85\linewidth]{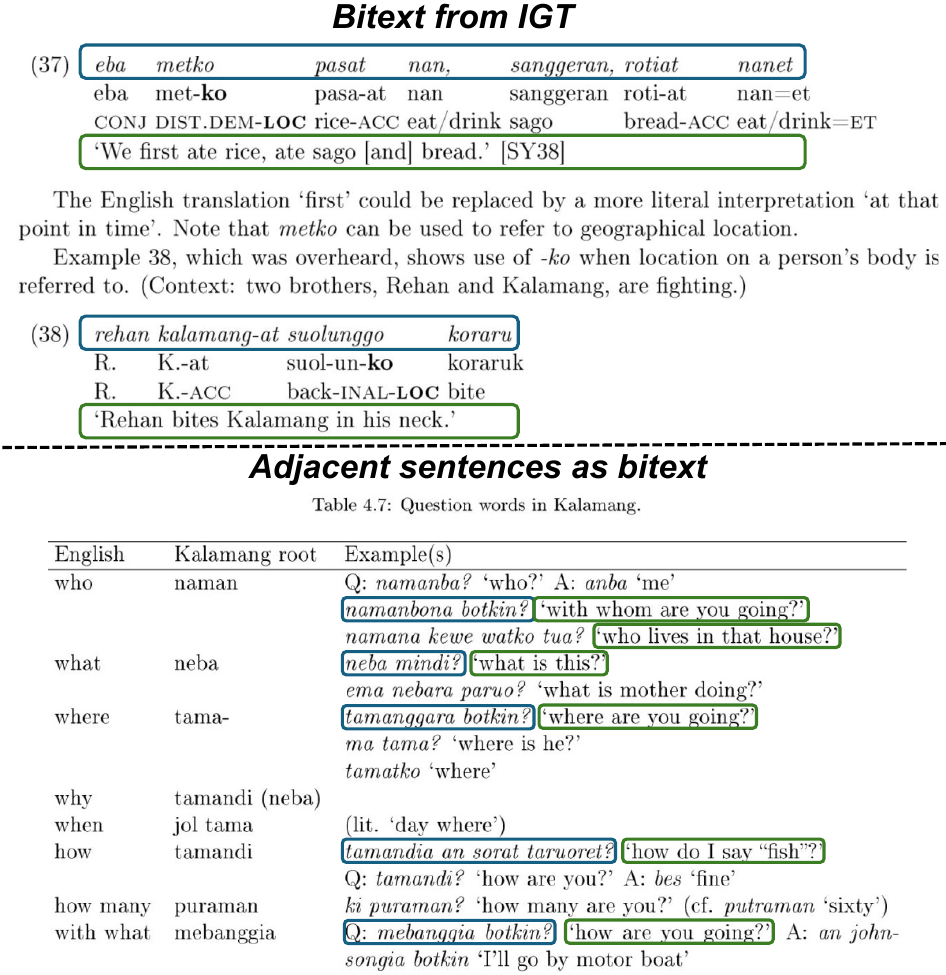}
    \vspace{-1em}
    \caption{Examples of incidental bitext in a Kalamang book~\cite{visser_kalamang2016}.
    Blue boxes indicate Kalamang sentences, and green boxes indicate English sentences.
    }
    \label{fig:ex_of_bitext}
\vspace{-1.5em}
\end{figure}

Therefore, in this work, we develop an (almost) fully automatic pipeline for extracting bitexts from heterogeneous linguistic resources in PDF format from the DReaM corpus.
Our pipeline consists of two stages.
First, we apply optical character recognition~(OCR) to recover text.
Second, we use a vision language model~(VLM) to identify parallel sentences using both the OCR output and page image.
Visual information is important because linguistic resources often encode parallelism through visual structure that is lost in plain text, such as IGT, adjacent translations, and parallel columns.
Figure~\ref{fig:ex_of_bitext} shows an example of such incidental bitext.
Applying our pipeline to 20 languages, we extract an average of 2,578 bitext examples per language.

\section{Related Work}

We review how previous work extracts structured data from linguistic resources.
Appendix~\ref{sec:bitext_alignment_methods} provides additional discussion of existing bitext alignment methods.

\subsection{Linguistic Resources Based MT}

Recent work has shown that grammar books and other linguistic resources can improve MT for \textit{unseen} languages when used with LLMs~\cite{tanzer2024a}.
A key challenge, however, is converting heterogeneous and poorly structured archival materials into usable structured data.

Several studies leverage structured or semi-structured machine readable resources.
\citet{tanzer2024a} use {\LaTeX} and plaintext grammar books together with well-formatted parallel sentences from \textit{Dictionaria}, while \citet{aycock2025can} extract parallel examples based on text formatting.

OCR is frequently used for archival materials.
Both \citet{zhang-etal-2024-hire} and \citet{hus-anastasopoulos-2024-back} use OCRed grammar books.
\citet{merx-etal-2024-low} OCR Mambai documents, then extract dictionaries based on text formatting and identify parallel sentences using Hunalign~\cite{varga-2007-parallel}. 

Manual intervention is sometimes needed.
\citet{zhang-etal-2025-read}, for example, manually extract grammar rules and parallel examples from a grammar book.
In some cases, extraction details are unspecified.
\citet{pei-etal-2025-understanding} extract parallel sentences from a dictionary without describing the procedure.

Unlike this line of prior work which primarily studies how different linguistic resources and prompting strategies impact MT performance, we focus on resource construction itself. 
We propose to extract useful data from archival linguistic resources with minimal manual intervention.

Most closely related to our work, \citet{setiawan2026contextvolumedrivesperformance} combine OCR with an LLM to extract parallel sentences and lexicons from PDF grammar books, but provide limited methodological detail and evaluate only one language.
\citet{setiawan2026mudiditwostageframeworkmultilingual} adopt a two-stage approach to leverage OCR and VLM for dictionary entry extraction, but do not consider bitext extraction or downstream MT evaluation.

\begin{table*}[t]
\small
    \centering
    \begin{tabular}{llllll}
    \toprule
    \multirow{2}{*}{\textbf{LRL name}} & \multirow{2}{*}{\textbf{LRL code}} & \multirow{2}{*}{\textbf{HRL}} & \multirow{2}{*}{\textbf{NLLB support}} & \multicolumn{2}{c}{\textbf{Existing data}}  \\
    \cmidrule{5-6}
    & & & & \textbf{Data source} & \textbf{Script} \\
    \midrule
    Arapaho & \texttt{arp} & \texttt{eng} & No & SIGMORPHON 2023 & Latin \\
    Chuvash & \texttt{chv} & \texttt{eng} & No & FLORES, BOUQuET & Cyrillic \\
    Tsez & \texttt{ddo} & \texttt{eng} & No & SIGMORPHON 2023 & Latin \\
    Dogri & \texttt{dgo} & \texttt{eng} & No & FLORES, BOUQuET & Devanagari \\
    Gitxsan & \texttt{git} & \texttt{eng} & No & SIGMORPHON 2023 & Latin \\
    Kalamang & \texttt{kgv} & \texttt{eng} & No & \citet{tanzer2024a} & Latin \\
    Lezghian & \texttt{lez} & \texttt{eng} & No & SIGMORPHON 2023 & Cyrillic \\
    Mambae & \texttt{mgm} & \texttt{eng} & No & \citet{merx-etal-2024-low} & Latin \\
    Manchu & \texttt{mnc} & \texttt{eng} & No & \citet{zhang-etal-2024-hire} & Latin \\
    Natügu & \texttt{ntu} & \texttt{eng} & No & SIGMORPHON 2023 & Latin \\
    Uspanteco & \texttt{usp} & \texttt{spa} & No & SIGMORPHON 2023 & Latin \\
    \midrule
    Southwestern Dinka & \texttt{dik} & \texttt{eng} & Yes & FLORES, BOUQuET & Latin \\
    Paraguayan Guaraní & \texttt{gug} & \texttt{eng} & Yes & FLORES, BOUQuET & Latin \\
    Iloko & \texttt{ilo} & \texttt{eng} & Yes & FLORES, BOUQuET & Latin \\
    Southern Jinghpaw & \texttt{kac} & \texttt{eng} & Yes & FLORES, BOUQuET & Latin \\
    Kabuverdianu & \texttt{kea} & \texttt{eng} & Yes & FLORES, BOUQuET & Latin \\
    Kimbundu & \texttt{kmb} & \texttt{por} & Yes & FLORES, BOUQuET & Latin \\
    Mizo & \texttt{lus} & \texttt{eng} & Yes & FLORES & Latin \\
    Minangkabau & \texttt{min} & \texttt{eng} & Yes & FLORES & Latin \\
    Wolof & \texttt{wol} & \texttt{eng} & Yes & FLORES, BOUQuET & Latin \\
    \bottomrule
    \end{tabular}
    \vspace{-1em}
    \caption{Summary of languages and their resource status, separated by NLLB support.
    Kimbundu is paired with Portuguese, for which our collection contains substantially more books, while Uspanteco is paired with Spanish to match its available test set.
    BOUQuET is only used as a secondary evaluation set where available.
    }
    \label{tab:langs_sum}
\end{table*}

\section{Languages and Status}
We select extremely under-resourced languages studied in prior work, which allows us to evaluate our method using existing test sets.
Table~\ref{tab:langs_sum} summarizes the languages and their resource status.

Our experimental design distinguishes between languages with and without web presence.
Languages supported by NLLB~\cite{nllbteam2022languageleftbehindscaling} have at least some, if not abundant, monolingual text online, making it likely that frontier LLMs have encountered them during pre-training.
In contrast, other languages can be treated as approximating languages that are unlikely to have been seen in LLM pre-training.
The latter group of languages allows us to simulate a setting in which a model is asked to translate a previously unseen language and the only available translation examples are extracted from archival resources.

\section{Method}
We extract and clean up bitexts from multilingual archival PDFs in three steps.
First, we apply multilingual OCR to recover text (Section~\ref{sec:ocr}).
Second, we use a VLM to identify bitext based on visual cues (Section~\ref{sec:vlm_align}).
Finally, we clean and filter the resulting alignments (Section~\ref{sec:post_align}).

\begin{table}[t]
\small
    \centering
    \begin{tabular}{lr}
    \toprule
    \textbf{Model} & \textbf{CER (\%)$\downarrow$} \\
    \midrule
    {\paddle} & 34.13 \\
    {\hunyuan} & 11.37 \\
    {\dotsocr} & 7.88 \\
    \midrule
    {\dotsocr} with retry & \textbf{5.66} \\
    \bottomrule
    \end{tabular}
    \vspace{-1em}
    \caption{CER for three OCR models on the synthesized FLORES PDFs.}
    \label{tab:ocr_benchmark}
\vspace{-1.5em}
\end{table}

\subsection{OCR}
\label{sec:ocr}

Recently developed neural OCR models outperform traditional systems in accuracy and human-readability~\cite{poznanski2025olmocr}, but their multilingual performance had not been comprehensively benchmarked when we began this work.
We therefore evaluate 3 multilingual models, {\dotsocr}~\cite{li2025dotsocrmultilingualdocumentlayout}, {\hunyuan}~\cite{hunyuanvisionteam2025hunyuanocrtechnicalreport}, and {\paddle}~\cite{cui2025paddleocrvlboostingmultilingualdocument} on synthetic multilingual PDFs.

Inspired by~\citet{ignat-etal-2022-ocr}, we render FLORES~\cite{nllbteam2022languageleftbehindscaling} devtest sentences into PDFs using \texttt{playwright}.\footnote{\url{https://github.com/microsoft/playwright}}
FLORES covers over 200 languages across 31 scripts.
We report macro-averaged character error rate (CER) across languages in Table~\ref{tab:ocr_benchmark}, with per-script results in Figure~\ref{fig:script_level_ocr_cer} in Appendix~\ref{sec:more_ocr}.

{\dotsocr} performs best and supports the broadest range of scripts, consistent with recent multilingual OCR benchmarks~\cite{kargaran2026glotocrbenchocrmodels,xu2026more}.
However, it occasionally produces endless repetitions.
Following~\citet{poznanski2025olmocr}, we retry inference with a higher temperature when generation exceeds the length limit, reducing CER from 7.88\% to 5.66\%.

\subsection{Vision Language Model}
\label{sec:vlm_align}

Incidental bitexts often exhibit certain \textit{visual properties}, such as IGT or adjacent sentences in different languages (Figure~\ref{fig:ex_of_bitext}).
We therefore use a VLM that takes a page image and its OCR text as input and identifies candidate bitext pairs based on visual cues.
Providing OCR text (1) avoids requiring the VLM to perform OCR, (2) leverages specialized OCR models that perform better on low-resource languages, and (3) enables the VLM to understand the content directly based on text.

We assume that each book primarily contains two languages: a low-resource language~(LRL) described in a high-resource language~(HRL).
The LRL is obtained from the resource metadata.
We identify the HRL by applying language identification~(langID) to each book, counting sentences per language, and selecting the most frequent language.
Since we target specific language directions (e.g., \verb|arp-eng|), we retain only books matching the target pair.
The VLM is expected to have substantial knowledge of the HRL and treat the remaining text as the LRL.

We use \texttt{Qwen3.5-27B-FP8}~\cite{qwen3.5} as the backbone VLM, served using \verb|vllm|~\cite{kwon2023efficient}.
Larger models performed substantially better in preliminary experiments, but computational constraints prevented us from using an even larger model.
Prompts are provided in Appendix~\ref{sec:prompts}.

\subsection{Postprocessing}
\label{sec:post_align}

Raw extracted sentence pairs are noisy.
For example, IGT may contain linguistic annotations, dictionaries may provide word-level rather than sentence-level alignments, and the VLM may produce incorrect alignments.
We therefore follow \citet{koehn-etal-2018-findings,koehn-etal-2019-findings} and apply language-agnostic cleaning and filtering (Section~\ref{sec:lang_all_filter}).
We additionally reconcile orthographic mismatches between archival resources and existing data using language-specific normalization (Section~\ref{sec:ortho_norm}).

We do not consider such postprocessing as a disadvantage, as conventional bitext mining pipelines likewise require extensive cleaning and filtering.

\subsubsection{Language-agnostic Cleaning and Filtering}
\label{sec:lang_all_filter}

We apply the following language-agnostic cleaning and filtering steps:
\begin{itemize}[leftmargin=0.5em, itemsep=0pt]
    
    \item Remove IGT annotation symbols (e.g., \verb|-|, \verb|=|, and \verb|ø|) from LRL sentences.
    
    \item Remove pairs where either sentence contains only digits or punctuation.

    \item Remove pairs where one sentence is more than twice as long as the other in characters.

    \item Remove pairs where neither sentence contains more than two space-separated words.

    \item Remove pairs with source-target Levenshtein similarity ratio above 0.9.

    \item Apply langID to both sentences. The HRL sentence must be identified as the HRL, while the LRL sentence must not be.
\end{itemize}

\subsubsection{Language-specific Orthography Normalization}
\label{sec:ortho_norm}

Orthographic mismatch, particularly in diacritics and script, is a major challenge in the extremely low-resource settings we are studying. Oftentimes, this is because of genuine diachronic changes in the orthography preferred by language communities and other stakeholders (e.g. documentary linguists) at different times.
For all languages, we apply minor corrections for common OCR errors, identified through manual inspection of sampled book pages, plus the following language-specific normalization.

\paragraph{arp, ilo, kac, mgm, mnc}
We remove diacritics.

\paragraph{chv}
We transliterate the Latin script used in our resources to the Cyrillic script used by FLORES, following an online document.\footnote{\url{https://assets.publishing.service.gov.uk/government/uploads/system/uploads/attachment_data/file/835784/ROMANIZATION_OF_CHUVASH.pdf}}

\paragraph{dgo}
Our resources primarily use the International Phonetic Alphabet (IPA), whereas FLORES uses Devanagari.
As we could not find existing conversion schemes, we let the coding agent Codex derive rules by comparing FLORES dev sentences with text from the archival resources.

\paragraph{git}
We remove diacritics.
The SIGMORPHON data additionally uses Combining Low Line (U+0332), e.g., \underline{K} (K followed by U+0332), which our OCR model fails to recognize.
We do not restore it.

\paragraph{lez}
We convert the Latin script used in our resources to the Cyrillic script used by SIGMORPHON, following \citet{Haspelmath+1993}.

\paragraph{ntu}
We convert the older orthography used in archival resources to the modern SIGMORPHON orthography following an online resource.\footnote{\url{https://en.wikipedia.org/wiki/Nat\%C3\%BCgu_language\#Orthography}}

Table~\ref{tab:lang_ortho_norm} in Appendix~\ref{sec:ortho_norm_detail} summarizes these procedures.
While the reader may be disappointed to see that such manual efforts are still needed in an otherwise fully automatic pipeline, these rules primarily reconcile archival orthographic conventions with modern usage and evaluation standards, rather than fixing bitext extraction itself. We also note that such manual verification processes might be needed regardless, to reconcile archival material orthographies with current community practices.

\begin{table}[t]
\small
    \centering
    \begin{tabular}{l|rr}
    \toprule
    \multirow{2}{*}{\textbf{Lang}} & \multicolumn{2}{c}{\textbf{BLASER 2.0-QE$\uparrow$}} \\
    \cmidrule{2-3}
    & \textbf{Extracted} & \textbf{Test / FLORES} \\
    \midrule
    \verb|dik| & 2.935 & \textbf{3.281} \\
    \verb|gug| & 3.375 & \textbf{3.732} \\
    \verb|ilo| & 3.356 & \textbf{4.274} \\
    \verb|kac| & 3.129 & \textbf{3.570} \\
    \verb|kea| & 3.293 & \textbf{3.965} \\
    \verb|kmb| & 2.470 & \textbf{3.344} \\
    \verb|lus| & 2.953 & \textbf{3.696} \\
    \verb|min| & 3.245 & \textbf{4.138} \\
    \verb|wol| & 3.123 & \textbf{3.511} \\
    \midrule
    \textbf{Avg}. & 3.098 & \textbf{3.723} \\
    \bottomrule
    \end{tabular}
    \vspace{-1em}
    \caption{
    QE scores for supported languages on extracted and FLORES bitexts.
    \textbf{Bold} indicates better performance.
    FLORES bitexts are of higher quality.
    }
    \label{tab:qe_scores}
\end{table}

\subsection{Intrinsic Evaluation}

We evaluate bitext quality using BLASER 2.0-QE~\cite{dale-costa-jussa-2024-blaser}, due to its broad language coverage.
For supported languages, we score each sentence pair and report average QE scores for the extracted bitext and corresponding test set in Table~\ref{tab:qe_scores}.
Scores for unsupported languages are unreliable and reported separately in Table~\ref{tab:qe_scores_unsupported} in the Appendix~\ref{sec:blaser_unsupported}.

\paragraph{Results}
The extracted bitexts achieve an average score of 3.098, compared with 3.723 for the test set.
Although this gap is substantial, a score of 3 indicates that the sentences are \textit{generally equivalent} in meaning, with only minor differences permitted under the XSTS scale~\cite{licht-etal-2022-consistent} predicted by BLASER 2.0-QE.

\begin{table}[t]
\small
    \centering
    \begin{tabular}{l|r|rrr}
    \toprule
    \multirow{2}{*}{\textbf{Lang}} & \multirow{2}{*}{\textbf{\#Extracted}} & \multicolumn{3}{c}{ \textbf{Existing Datasets}} \\
    \cmidrule{3-5}
    & & \#Train & \#Dev & \#Test \\
    \midrule
    \multicolumn{5}{l}{\textit{Without NLLB support}} \\
    \texttt{arp} & 1,796 & 36,123 & 4,892 & 4,892 \\
    \texttt{chv} & 454 & -- & 997 / 504 & 1,012 / 854 \\
    \texttt{ddo} & 1,403 & 3,547 & 445 & 445 \\
    \texttt{dgo} & 137 & -- & 498 / 504 & 499 / 854 \\
    \texttt{git} & 1,018 & 31 & 42 & 37 \\
    \texttt{kgv} & 1,545 & 375 & -- & 100 \\
    \texttt{lez} & 1,533 & 634 & 88 & 87 \\
    \texttt{mgm} & 1,253 & 1,187 & -- & 50 \\
    \texttt{mnc} & 953 & 3 & -- & 102 \\
    \texttt{ntu} & 1,118 & 789 & 99 & 99 \\
    \texttt{usp} & 1,007 & 9,116 & 232 & 633 \\
    \midrule
    \multicolumn{5}{l}{\textit{With NLLB support}} \\
    \texttt{dik} & 3,238 & -- & 997 / 504 & 1,012 / 854 \\
    \texttt{gug} & 910 & -- & 997 / 504 & 1,012 / 854 \\
    \texttt{ilo} & 14,263 & -- & 997 / 504 & 1,012 / 854 \\
    \texttt{kac} & 817 & -- & 997 / 504 & 1,012 / 854 \\
    \texttt{kea} & 1,483 & -- & 997 / 504 & 1,012 / 854 \\
    \texttt{kmb} & 13,067 & -- & 997 / 504 & 1,012 / 854 \\
    \texttt{lus} & 2,477 & -- & 997 & 1,012 \\
    \texttt{min} & 459 & -- & 997 & 1,012 \\
    \texttt{wol} & 2,632 &  -- & 997 / 504 & 1,012 / 854 \\
    \bottomrule
    \end{tabular}
    \vspace{-1em}
    \caption{
    Number of sentences in the extracted and existing datasets.
    For FLORES, the devtest set is used as the test set.
    In cells with two numbers, the left one corresponds to FLORES and the right one to BOUQuET.
    }
    \label{tab:data_count}
\vspace{-.5em}
\end{table}

\section{MT Experiments for Extrinsic Evaluation}
We further assess the quality of extracted bitexts through few-shot LLM translation and MT model fine-tuning.

\subsection{Training and Evaluation Data}
As in Table~\ref{tab:langs_sum}, 11 languages are evaluated by FLORES,\footnote{\url{https://huggingface.co/datasets/openlanguagedata/flores_plus}} 6 are evaluated using the SIGMORPHON 2023 Interlinear Glossing shared task data~\cite{ginn-etal-2023-findings},\footnote{\url{https://github.com/sigmorphon/2023glossingST}} and the remaining 3 rely on test sentences curated in prior work.
We additionally use BOUQuET~\cite{andrews-etal-2025-bouquet} as a \textit{secondary} evaluation set for the 9 supported languages.

We detokenize the SIGMORPHON test sentences using Moses~\cite{koehn-etal-2007-moses}, while leaving other test sets unchanged.

To prevent train-test overlap, we remove extracted sentence pairs whose source or target has Levenshtein similarity ratio above 0.9 with any dev or test sentence.

Table~\ref{tab:data_count} reports the number of extracted sentences after filtering and the sizes of existing datasets.

\subsection{Experiment Setup}

\paragraph{Few-shot experiments}
We use \texttt{Qwen3.6-27B}~\cite{qwen3.6-27b} (hereafter \texttt{Qwen}).
Additional results with the smaller \texttt{Llama-3.1-8B}~\cite{grattafiori2024llama3herdmodels} are reported in Appendix~\ref{sec:few_shot_llama}.
Prompt templates are provided in Appendix~\ref{sec:prompts}.

For example retrieval, we score candidate sentences with the input sentence by chrF2++~\cite{popovic-2017-chrf} and select the highest-scoring pairs.
Prior work finds that retrieval method has limited impact on the peak MT performance~\cite{setiawan2026contextvolumedrivesperformance}.
An ablation study in Appendix~\ref{sec:few_shot_ablation}, though, shows that 512 examples perform best.

\paragraph{Fine-tuning experiments}
We fine-tune \texttt{nllb-200-distilled-1.3B} (hereafter \texttt{nllb}) using LoRA adapters~\cite{hu2022lora}.
Due to the tiny datasets, we train one multilingual model on all extracted data using temperature-based sampling.
More details are in Appendix~\ref{sec:nllb_finetune}.

\paragraph{Baselines}
For languages supported by NLLB, we compare the fine-tuned \texttt{nllb} against the out-of-the-box model. 
For unsupported LRL-to-HRL translaion, we use the proxy source language codes in Table~\ref{tab:xsim}.\footnote{In some cases, proxy source and target codes are identical (e.g., \texttt{eng\_Latn} for both sides of \texttt{arp-eng}). This has negligible impact, as alternative proxies yield similar results.}
For all languages, we compare few-shot translation using \texttt{Qwen} against zero-shot.

\paragraph{Evaluation Metrics}

To evaluate, we lowercase hypotheses and references and remove punctuation except apostrophes.
We report spBLEU\footnote{nrefs:1|case:lc|eff:no|tok:flores200|smooth:exp|version:2.6.0} with the flores200 tokenizer~\cite{nllbteam2022languageleftbehindscaling} and chrF2++,\footnote{nrefs:1|case:lc|eff:yes|nc:6|nw:2|space:no|version:2.6.0}  both computed with SacreBLEU~\cite{post-2018-call}.
We compute statistical significance using paired bootstrap resampling~\cite{koehn-2004-statistical}.\footnote{bs:1000|seed:12345}

\begin{table}[t]
\footnotesize
    \centering
    \begin{tabular}{l|rr|rr}
    \toprule
    \multirow{2}{*}{\textbf{Lang}} & \multicolumn{2}{c|}{\texttt{Qwen}} & \multicolumn{2}{c}{\texttt{nllb}} \\
    \cmidrule(lr){2-3} \cmidrule(lr){4-5} 
    & \multicolumn{1}{c}{0 shot} & \multicolumn{1}{c}{512 shot} & \multicolumn{1}{c}{OOB} & \multicolumn{1}{c}{fine-tuned} \\
    \midrule
\multicolumn{5}{l}{\textit{Without NLLB support}} \\
\verb|arp| & 3.86 & \textsuperscript{*}\upcell{32.90}{29.04} & -- & 9.96 \\
\verb|chv| & 20.99 & \textsuperscript{*}\upcell{30.49}{9.50} & -- & 5.14 \\
\verb|ddo| & 0.17 & \textsuperscript{*}\upcell{24.24}{24.07} & -- & 1.93 \\
\verb|dgo| & 21.18 & \textsuperscript{*}\downcell{20.68}{0.50} & -- & 4.12 \\
\verb|git| & 5.50 & \textsuperscript{*}\upcell{23.16}{17.66} & -- & 3.56 \\
\verb|kgv| & 8.38 & \textsuperscript{*}\upcell{39.23}{30.85} & -- & 7.60 \\
\verb|lez| & 9.58 & \textsuperscript{*}\upcell{25.26}{15.68} & -- & 3.97 \\
\verb|mgm| & 17.34 & \textsuperscript{*}\upcell{37.91}{20.57} & -- & 18.78 \\
\verb|mnc| & 3.92 & \textsuperscript{*}\upcell{27.65}{23.73} & -- & 3.03 \\
\verb|ntu| & 5.51 & \textsuperscript{*}\upcell{48.76}{43.25} & -- & 9.39 \\
\verb|usp| & 6.51 & \textsuperscript{*}\upcell{28.09}{21.58} & -- & 7.15 \\
\textit{Avg}. & \textit{9.36} & \upcell{\textit{30.76}}{21.40} & -- & \textit{6.78} \\
\midrule
\multicolumn{5}{l}{\textit{With NLLB support}} \\
\verb|dik| & 6.36 & \textsuperscript{*}\upcell{16.89}{10.53} & 19.81 & \textsuperscript{*}\downcell{11.28}{8.53} \\
\verb|gug| & 14.90 & \textsuperscript{*}\upcell{26.05}{11.15} & 37.27 & \textsuperscript{*}\downcell{31.73}{5.54} \\
\verb|ilo| & 42.39 & \textsuperscript{*}\upcell{46.76}{4.37} & 52.56 & \textsuperscript{*}\downcell{27.83}{24.73} \\
\verb|kac| & 4.01 & \textsuperscript{*}\upcell{25.91}{21.90} & 38.56 & \textsuperscript{*}\downcell{26.03}{12.53} \\
\verb|kea| & 29.50 & \textsuperscript{*}\upcell{40.26}{10.76} & 45.33 & \textsuperscript{*}\downcell{39.37}{5.96} \\
\verb|kmb| & 12.33 & \textsuperscript{*}\upcell{16.19}{3.86} & 24.96 & \textsuperscript{*}\downcell{10.46}{14.50} \\
\verb|lus| & 15.37 & \textsuperscript{*}\upcell{24.56}{9.19} & 36.70 & \textsuperscript{*}\downcell{8.73}{27.97} \\
\verb|min| & 39.36 & \textsuperscript{*}\upcell{43.40}{4.04} & 49.50 & \textsuperscript{*}\downcell{26.02}{23.48} \\
\verb|wol| & 22.62 & \textsuperscript{*}\upcell{26.93}{4.31} & 25.72 & \textsuperscript{*}\downcell{9.01}{16.71} \\
\textit{Avg}. & \textit{20.76} & \upcell{\textit{29.66}}{8.90} & \textit{36.71} & \downcell{\textit{21.16}}{15.55} \\
    \midrule
    \textbf{Avg}. & 14.49 & \upcell{30.27}{15.78} & -- & 13.25 \\
    \bottomrule
    \end{tabular}
    \vspace{-1em}
    \caption{
    Extracted data improves few-shot translation performance \textit{into LRLs}.
    Results are reported in chrF2++($\uparrow$).
    OOB denotes out-of-the-box performance.
    \textsuperscript{*}$\text{p-value} <0.05$.
    }
    \label{tab:into_lrl_mt}
\vspace{-1.5em}
\end{table}

\begin{table}[t]
\footnotesize
    \centering
    \begin{tabular}{l|rr|rr}
    \toprule
    \multirow{2}{*}{\textbf{Lang}} & \multicolumn{2}{c|}{\texttt{Qwen}} & \multicolumn{2}{c}{\texttt{nllb}} \\
    \cmidrule(lr){2-3} \cmidrule(lr){4-5} 
    & \multicolumn{1}{c}{0 shot} & \multicolumn{1}{c}{512 shot} & \multicolumn{1}{c}{OOB} & \multicolumn{1}{c}{fine-tuned} \\
    \midrule
\multicolumn{5}{l}{\textit{Without NLLB support}$^\dagger$} \\
\verb|arp| & 11.67 & \textsuperscript{*}\upcell{22.09}{10.42} & 8.62 & \textsuperscript{*}\upcell{12.59}{3.97} \\
\verb|chv| & 47.46 & \textsuperscript{*}\upcell{48.66}{1.20} & 19.72 & \textsuperscript{*}\upcell{24.47}{4.75} \\
\verb|ddo| & 14.76 & \textsuperscript{*}\upcell{27.64}{12.88} & 10.27 & \textsuperscript{*}\upcell{16.55}{6.28} \\
\verb|dgo| & 47.52 & \textsuperscript{*}\upcell{50.50}{2.98} & 42.86 & \textsuperscript{*}\downcell{39.53}{3.33} \\
\verb|git| & 25.39 & \textsuperscript{*}\upcell{30.49}{5.10} & 15.42 & \textsuperscript{*}\upcell{24.87}{9.45} \\
\verb|kgv| & 13.30 & \textsuperscript{*}\upcell{33.38}{20.08} & 11.89 & \textsuperscript{*}\upcell{18.04}{6.15} \\
\verb|lez| & 24.85 & \textsuperscript{*}\upcell{34.51}{9.66} & 9.64 & \textsuperscript{*}\upcell{20.51}{10.87} \\
\verb|mgm| & 30.11 & \textsuperscript{*}\upcell{42.99}{12.88} & 20.61 & \textsuperscript{*}\upcell{30.30}{9.69} \\
\verb|mnc| & 18.27 & \textsuperscript{*}\upcell{24.85}{6.58} & 8.51 & \textsuperscript{*}\upcell{12.33}{3.82} \\
\verb|ntu| & 15.23 & \textsuperscript{*}\upcell{40.01}{24.78} & 9.44 & \textsuperscript{*}\upcell{38.73}{29.29} \\
\verb|usp| & 11.74 & \textsuperscript{*}\upcell{26.05}{14.31} & 9.98 & \textsuperscript{*}\upcell{16.07}{6.09} \\
\textit{Avg}. & \textit{23.66} & \upcell{\textit{34.65}}{10.99} & \textit{15.18} & \upcell{\textit{23.09}}{7.91} \\
\midrule
\multicolumn{5}{l}{\textit{With NLLB support}} \\
\verb|dik| & 23.94 & \textsuperscript{*}\upcell{25.39}{1.45} & 29.49 & \textsuperscript{*}\downcell{28.99}{0.50} \\
\verb|gug| & 42.50 & \textsuperscript{*}\upcell{44.06}{1.56} & 47.96 & \textsuperscript{*}\downcell{45.19}{2.77} \\
\verb|ilo| & 56.33 & \textsuperscript{*}\upcell{58.02}{1.69} & 60.14 & \textsuperscript{*}\downcell{58.58}{1.56} \\
\verb|kac| & 24.24 & \textsuperscript{*}\upcell{26.83}{2.59} & 38.29 & \textsuperscript{*}\downcell{37.36}{0.93} \\
\verb|kea| & 64.93 & \textsuperscript{*}\upcell{69.59}{4.66} & 67.64 & \downcell{67.19}{0.45} \\
\verb|kmb| & 24.96 & \textsuperscript{*}\upcell{25.42}{0.46} & 30.62 & \textsuperscript{*}\downcell{29.89}{0.73} \\
\verb|lus| & 34.05 & \textsuperscript{*}\upcell{35.83}{1.78} & 41.21 & \textsuperscript{*}\downcell{39.09}{2.12} \\
\verb|min| & 58.25 & \textsuperscript{*}\upcell{60.40}{2.15} & 60.18 & \textsuperscript{*}\downcell{56.66}{3.52} \\
\verb|wol| & 38.14 & \textsuperscript{*}\upcell{39.75}{1.61} & 39.21 & \textsuperscript{*}\downcell{36.73}{2.48} \\
\textit{Avg}. & \textit{40.82} & \upcell{\textit{42.81}}{1.99} & \textit{46.08} & \downcell{\textit{44.41}}{1.67} \\
    \midrule
    \textbf{Avg.} & 31.38 & \upcell{38.32}{6.94} & 29.09 & \upcell{32.68}{3.59} \\
    \bottomrule
    \end{tabular}
    \vspace{-1em}
    \caption{
    Extracted data improves few-shot translation performance \textit{into HRLs}.
    Results are reported in chrF2++($\uparrow$). 
    $^\dagger$Proxy source language codes are used.
    \textsuperscript{*}$\text{p-value}<0.05$.
    }
    \label{tab:into_hrl_mt}
\vspace{-1em}
\end{table}

\section{Results and Analysis}

We present main results in Section~\ref{sec:main_results}, analyze the limitations of traditional sentence alignment methods in Section~\ref{sec:traditional_sent_mining}, discuss the limitations of archival resource utility in Section~\ref{sec:gold_few_shot}, and explore broader applicability in Section~\ref{sec:creolemt}.

\subsection{Main Results}
\label{sec:main_results}

Tables~\ref{tab:into_lrl_mt} and~\ref{tab:into_hrl_mt} report chrF2++ scores for translation in and out of LRLs respectively.
Corresponding spBLEU results are in Appendix Tables~\ref{tab:into_lrl_mt_spbleu} and~\ref{tab:into_hrl_mt_spbleu}.
Appendix~\ref{sec:bouquet_result} reports results on~BOUQuET.

Extracted bitexts substantially improve \texttt{Qwen} performance for all languages except \texttt{dgo}, with average gains of 15.78 and 6.94 chrF2++ for translation into LRLs and HRLs, respectively.
The lack of improvement for \texttt{dgo} is likely due to its automatically derived orthography normalization rules.

Fine-tuning \texttt{nllb} does not improve translation into LRLs, likely because of the limited training data~\cite{koehn-knowles-2017-six}.
However, it improves translation into HRLs by 3.59 chrF2++ on average.
Overall, we show that the extracted bitexts provide useful supervision even at this small scale.

\subsection{A Failed Attempt: Bitext Extraction with Existing Sentence Alignment Methods}
\label{sec:traditional_sent_mining}

Given OCRed text, we also attempted to mine bitext using established sentence alignment methods.
However, they perform poorly.

\begin{table}[t]
\small
    \centering
    \begin{tabular}{l|r|r|lr}
    \toprule
    \multirow{2}{*}{\textbf{Lang}}  & \textbf{LASER2} & \textbf{LABSE} & \multicolumn{2}{c}{\textbf{SONAR}} \\
    & \multicolumn{1}{c|}{xsim$\downarrow$} & \multicolumn{1}{c|}{xsim$\downarrow$} & \multicolumn{1}{c}{lang code} & \multicolumn{1}{c}{xsim$\downarrow$} \\
    \midrule
\multicolumn{5}{l}{\textit{Without NLLB support}$^\dagger$} \\
    \verb|arp| & 96.89 & \textbf{94.60} & \verb|eng_Latn| & 95.34 \\
    \verb|chv| & 76.19 & 33.70 & \verb|tat_Cyrl| & \textbf{33.40} \\
    \verb|ddo| & 97.08 & \textbf{90.34} & \verb|eng_Latn| & 90.79 \\
    \verb|dgo| & 40.48 & 0.80 & \verb|hin_Deva| & \textbf{0.20} \\
    \verb|git| & 62.16 & \textbf{43.24} & \verb|eng_Latn| & 54.05 \\
    \verb|kgv| & 84.00 & 61.00 & \verb|ind_Latn| & \textbf{52.00} \\
    \verb|lez| & 87.36 & \textbf{51.72} & \verb|rus_Cyrl| & 77.01 \\
    \verb|mgm| & 46.00 & \textbf{28.00} & \verb|ind_Latn| & 34.00 \\
    \verb|mnc| & \textbf{93.14} & 95.10 & \verb|khk_Cyrl| & \textbf{93.14} \\
    \verb|ntu| & 80.81 & \textbf{52.53} & \verb|ind_Latn| & 70.71 \\
    \verb|usp| & 96.37 & 95.58 & \verb|spa_Latn| & \textbf{94.79} \\
\midrule
\multicolumn{5}{l}{\textit{With NLLB support}} \\
    \verb|dik| & 79.45 & 38.04 & \verb|dik_Latn| & \textbf{9.58} \\
    \verb|gug| & 71.74 & 17.09 & \verb|grn_Latn| & \textbf{0.10} \\
    \verb|ilo| & 61.96 & 6.62 & \verb|ilo_Latn| & \textbf{0.00} \\
    \verb|kac| & 91.50 & 40.32 & \verb|kac_Latn| & \textbf{0.20} \\
    \verb|kea| & 22.04 & 0.49 & \verb|kea_Latn| & \textbf{0.00} \\
    \verb|kmb| & 74.01 & 39.62 & \verb|kmb_Latn| & \textbf{0.89} \\
    \verb|lus| & 87.15 & 23.22 & \verb|lus_Latn| & \textbf{0.49} \\
    \verb|min| & 42.59 & 1.19 & \verb|min_Latn| & \textbf{0.00} \\
    \verb|wol| & 76.38 & 26.19 & \verb|wol_Latn| & \textbf{0.69} \\
    \midrule
    \textbf{Avg.} & 73.37 & 41.97 & \multicolumn{2}{r}{\textbf{35.37}} \\
    \bottomrule
    \end{tabular}
    \vspace{-1em}
    \caption{xSIM error rate (\%) on test sets.
    SONAR achieves the lowest error rate.
    $^\dagger$Proxy SONAR language codes are used.
    }
    \label{tab:xsim}
\end{table}

\paragraph{Sentence Embedding Evaluation}

State-of-the-art sentence alignment methods such as Global Mining~\cite{schwenk-etal-2021-ccmatrix} and Vecalign~\cite{thompson-koehn-2019-vecalign} depend on high-quality sentence embeddings.
We therefore evaluate embedding quality for our low-resource languages using xSIM (multilingual similarity search).\footnote{\url{https://github.com/facebookresearch/LASER/blob/main/tasks/xsim/README.md}}
Given aligned bitext, xSIM embeds both source and target sentences into a shared space and retrieves the best target for each source sentence using a margin score~\cite{artetxe-schwenk-2019-margin}, where $x$ and $y$ are sentence embeddings and $\text{NN}_k(\cdot)$ finds the $k$ nearest neighbors:
\begin{align}
\label{eq:margin_score}
\begin{split}
&\text{sim}(x, y) \\
&= \frac{\text{cos}(x,y)}{\sum_{z \in \text{NN}_k(x)}\frac{\text{cos}(x,z)}{2k} + \sum_{z \in \text{NN}_k(y)}\frac{\text{cos}(y,z)}{2k}}
\end{split}
\end{align}

We report alignment error against gold alignments.
We evaluate LASER2~\cite{heffernan-etal-2022-bitext}, LABSE~\cite{feng-etal-2022-language}, and SONAR embeddings~\cite{duquenne2023sonar} on test sets listed in Table~\ref{tab:langs_sum}.
We use LASER2 rather than LASER3 because the latter is language-specific.
And for languages unsupported by SONAR, we use the closest supported language code.

Table~\ref{tab:xsim} reports xSIM error rates.
Although SONAR performs best overall, error rates remain high for unseen languages such as Mambai (34\%) and Kalamang (52\%), limiting the effectiveness of embedding-based alignment.

\begin{table}[t]
\small
    \centering
    \begin{tabular}{lrrr}
    \toprule
    \textbf{Alignment Method} &  \textbf{Precision} & \textbf{Recall} & \textbf{F1} \\
    \midrule
    \multicolumn{4}{l}{\textit{Methods without prior knowledge}} \\
    Global Mining & 0.0610 & 0.0549 & 0.0578 \\
    Local Mining & 0.0851 & 0.2473 & 0.1266 \\
    Vecalign & 0.2776 & 0.4560 & 0.3451 \\
    Hunalign w/o dict. & 0.2435 & 0.2582 & 0.2507 \\
    \textbf{VLM (our method)} & \textbf{1.0000} & \textbf{0.7657} & \textbf{0.8673} \\ 
    \midrule
    \multicolumn{4}{l}{\textit{Methods with prior knowledge}} \\
    Hunalign w/ dict. & 0.5876 & 0.6264 & 0.6064 \\
    \bottomrule
    \end{tabular}
    \vspace{-1em}
    \caption{
    Alignment performance of different methods.
    VLM achieves the best performance.
    Its imperfect recall is due to pages dominated by word-level alignments, where sentence-level alignments are overlooked.
    }
    \label{tab:align_acc}
    \vspace{-.5em}
\end{table}

\paragraph{Sentence Alignment Evaluation}

We evaluate alignment methods on \verb|mgm-eng|, which has a relatively low SONAR xSIM error rate.
Alignment performance on languages with higher xSIM error rates is expected to be worse.

We first segment OCR text into sentences, and use a langID model to separate two languages in the book.
We then apply embedding based \textit{Global Mining}, \textit{Local Mining}, \textit{Vecalign}, and non-embedding based \textit{Hunalign} to OCRed book content page by page.
Further details are provided in Appendix~\ref{sec:alignment_details}.

We manually align 182 sentence pairs from randomly sampled pages of a Mambae manual~\cite{sHullMambaiAinaro} as groundtruths and report precision, recall, and F1 in Table~\ref{tab:align_acc}.
Without an auxiliary dictionary, traditional alignment methods perform poorly, while VLM achieves highest F1. 
Beyond poor embedding quality, text preprocessing introduces substantial errors.
Sentence segmentation can fail when multiple parallel pairs occur on the same line or when OCR splits a single sentence across multiple lines.
LangID errors can cause sentences in the same language to be incorrectly aligned.

VLM-based alignment avoids these failure modes by directly accessing images. 
It can infer sentence boundaries and correspondences from visual structure without separate sentence splitting, langID, or auxiliary dictionaries, producing substantially better alignment.

\begin{table}[t]
\footnotesize
    \centering
    \begin{tabular}{l|rr|rr}
    \toprule
    \multirow{2}{*}{\textbf{Lang}} & \multicolumn{2}{c}{\textbf{Into LRL}} & \multicolumn{2}{c}{\textbf{Into HRL}} \\
    & \multicolumn{1}{c}{\textbf{Gold}} & \multicolumn{1}{c}{\textbf{Book}} & \multicolumn{1}{c}{\textbf{Gold}}  & \multicolumn{1}{c}{\textbf{Book}}  \\
    \cmidrule{1-1} \cmidrule(lr){2-3} \cmidrule(lr){4-5} 
\multicolumn{5}{l}{\textit{Without NLLB support}} \\
\verb|arp| & 48.32 & \textsuperscript{*}\downcell{32.90}{15.42} & 39.30 & \textsuperscript{*}\downcell{22.09}{17.21} \\
\verb|chv| & 33.77 & \textsuperscript{*}\downcell{30.49}{3.28} & 49.96 & \textsuperscript{*}\downcell{48.66}{1.30} \\
\verb|ddo| & 43.59 & \textsuperscript{*}\downcell{24.24}{19.35} & 42.38 & \textsuperscript{*}\downcell{27.64}{14.74} \\
\verb|dgo| & 34.52 & \textsuperscript{*}\downcell{20.68}{13.84} & 54.62 & \textsuperscript{*}\downcell{50.50}{4.12} \\
\verb|git| & 21.28 & \upcell{23.16}{1.88} & 26.08 & \textsuperscript{*}\upcell{30.49}{4.41} \\
\verb|kgv| & 38.97 & \upcell{39.23}{0.26} & 34.56 & \downcell{33.38}{1.18} \\
\verb|lez| & 32.13 & \textsuperscript{*}\downcell{25.26}{6.87} & 41.42 & \textsuperscript{*}\downcell{34.51}{6.91} \\
\verb|mgm| & 33.44 & \textsuperscript{*}\upcell{37.91}{4.47} & 42.08 & \upcell{42.99}{0.91} \\
$^\dagger$\verb|mnc| & 7.84 & \textsuperscript{*}\upcell{27.65}{19.81} & 20.67 & \textsuperscript{*}\upcell{24.85}{4.18} \\
\verb|ntu| & 47.79 & \upcell{48.76}{0.97} & 39.00 & \upcell{40.01}{1.01} \\
\verb|usp| & 42.15 & \textsuperscript{*}\downcell{28.09}{14.06} & 42.21 & \textsuperscript{*}\downcell{26.05}{16.16} \\
\midrule
\multicolumn{5}{l}{\textit{With NLLB support}} \\
\verb|dik| & 23.98 & \textsuperscript{*}\downcell{16.89}{7.09} & 27.09 & \textsuperscript{*}\downcell{25.39}{1.70} \\
\verb|gug| & 34.84 & \textsuperscript{*}\downcell{26.05}{8.79} & 45.50 & \textsuperscript{*}\downcell{44.06}{1.44} \\
\verb|ilo| & 51.30 & \textsuperscript{*}\downcell{46.76}{4.54} & 59.09 & \textsuperscript{*}\downcell{58.02}{1.07} \\
\verb|kac| & 39.84 & \textsuperscript{*}\downcell{25.91}{13.93} & 29.89 & \textsuperscript{*}\downcell{26.83}{3.06} \\
\verb|kea| & 57.46 & \textsuperscript{*}\downcell{40.26}{17.20} & 71.50 & \textsuperscript{*}\downcell{69.59}{1.91} \\
\verb|kmb| & 33.62 & \textsuperscript{*}\downcell{16.19}{17.43} & 30.32 & \textsuperscript{*}\downcell{25.42}{4.90} \\
\verb|lus| & 37.17 & \textsuperscript{*}\downcell{24.56}{12.61} & 38.54 & \textsuperscript{*}\downcell{35.83}{2.71} \\
\verb|min| & 51.12 & \textsuperscript{*}\downcell{43.40}{7.72} & 62.94 & \textsuperscript{*}\downcell{60.40}{2.54} \\
\verb|wol| & 32.07 & \textsuperscript{*}\downcell{26.93}{5.14} & 42.06 & \textsuperscript{*}\downcell{39.75}{2.31} \\
    \midrule
    \textbf{Avg.} & 38.81 & \downcell{30.40}{8.41} & 43.08 & \downcell{38.99}{4.09} \\
    \bottomrule
    \end{tabular}
    \vspace{-1em}
    \caption{
    chrF2++ scores for 512-shot translation with \texttt{Qwen}.
    Extracted bitexts are narrowing the gap to gold in-domain bitexts.
    $^\dagger$Excluded from the average as it has only 3 gold examples.
    \textsuperscript{*}$\text{p-value} <0.05$.
    }
    \label{tab:gold_few_shot}
    \vspace{-.5em}
\end{table}

\subsection{Using In-domain Parallel Sentences for Prompting}
\label{sec:gold_few_shot}

Section~\ref{sec:main_results} assumes no additional parallel examples and uses only potentially out-of-domain extracted bitexts.
As a skyline, we use available in-domain training or dev set sentences to quantify the gap between in-domain and extracted examples.

Table~\ref{tab:gold_few_shot} compares few-shot MT performance with book examples and gold in-domain examples.
We use FLORES dev set sentences when available and training set sentences otherwise.

Compared with book examples, in-domain examples improve the average chrF2++ from 30.40 to 38.81 for translation into LRLs and by 4.09 chrF2++ for into HRLs.
For most languages, in-domain examples substantially outperform book examples, likely due to their higher quality (Table~\ref{tab:qe_scores}).
Interestingly, book examples perform better in \verb|git|, \verb|mgm|, \verb|mnc|, and \verb|ntu|, possibly because more book examples are available and some books overlap with those used to construct the test sets.
It suggests that extracting bitexts from books is still a feasible way towards narrowing the gap in LRL translation when gold bitexts are unavailable.

\subsection{Broader Applicability: CreoleMT}
\label{sec:creolemt}

To examine broader applicability, we apply our pipeline to 27 of the lowest-resource Creole languages in Kreyòl-MT~\cite{robinson-etal-2024-kreyol}.
Unlike our main experiments, we use only language-agnostic cleaning and filtering, without language-specific orthographic normalization.
We evaluate few-shot experiments using \texttt{Qwen} on the public Kreyòl-MT test set using the same evaluation metrics. %
Results are reported in Table~\ref{tab:creole_mt} in Appendix~\ref{sec:creole_mt_results}.

Extracted bitexts substantially improve translation into Creole languages from 16.87 to 35.71 chrF2++ on average, and translation into English from 34.54 to 48.79 chrF2++.
These results suggest that the pipeline generalizes to a broader set of low-resource languages even without language-specific human intervention.

\section{Conclusion}

We introduce a vision language model based pipeline for extracting parallel sentences from archival linguistic resources by leveraging their visual structure.
Across more than 20 languages, few-shot translation with an LLM using the extracted bitext substantially improves low-resource MT over zero-shot translation, demonstrating the value of archival linguistic resources as a source of parallel data.

\section*{Limitations}
Orthography mismatches remain challenging for archival materials.
Proper normalization still requires manually designed language-specific rules, while OCR systems can struggle with diacritics and visually similar characters encoded as different Unicode code points (e.g., Latin and Cyrillic \verb|a|).

Not all archival books are equally suitable sources of parallel data.
Many of the books in our collection were published decades ago and may use historical or outdated orthographic conventions that differ from those used in contemporary evaluation data.
Moreover, given the heterogeneous nature of archival linguistic resources, some books primarily cover topics such as phonology or grammatical descriptions and may contain relatively little parallel text useful for machine translation.
Careful resource selection is therefore important for obtaining high-quality and relevant examples.
Future advances in OCR and agentic systems might help automate these steps and further reduce manual intervention.

\section*{Acknowledgments}
This material is partially supported by the National Science Foundation under CAREER award
2439202 and partially based upon work supported by the ASA(ALT) SBIR CcoE under Contract No. W51701-25-C-0060.
Any opinions, findings and conclusions or recommendations expressed in this material are those of the author(s) and do not necessarily reflect the views of the ASA(ALT) SBIR CcoE.
We acknowledge the support by resources provided by the Office of Research Computing at George Mason University (\url{https://orc.gmu.edu}) and funded in part by grants from the NSF (Award Number
2018631).

\bibliography{custom,anthology-1,anthology-2}

\appendix

\section{Related Work on Bitext Alignment}
\label{sec:bitext_alignment_methods}

Bitext alignment algorithms have been popular to create sentence-aligned bitext for training MT systems.
They can be broadly divided into bitext mining and sentence alignment.

Bitext mining~\cite{schwenk-etal-2021-wikimatrix, schwenk-etal-2021-ccmatrix,nllbteam2022languageleftbehindscaling} relies on language-agnostic multilingual sentence embedding models, such as LaBSE~\cite{feng-etal-2022-language}, LASER~\cite{artetxe-schwenk-2019-massively, heffernan-etal-2022-bitext}, and SONAR~\cite{duquenne2023sonar}, which map sentences of similar meanings to proximate points in a shared, fixed-size embedding space, regardless of the language.
In the mining process, documents collected from web are first split into sentences, and language identification (langID) is applied to group sentences by language.
Sentence embeddings are then computed for each sentence.
Margin-based embedding similarities~\cite{artetxe-schwenk-2019-margin} are computed for each pair of sentences, with the ones with highest similarities are used as bitext.
The resulting bitext are often noise and require extensive filtering to improve quality~\cite{koehn-etal-2019-findings,sloto-etal-2023-findings}.

Another line of work, sentence alignment, assumes the availability of parallel documents.
A pair of parallel documents contain the same content but differ in language.
Sentence alignment methods typically use dynamic programming with scoring functions based on sentence lengths~\cite{brown-etal-1991-aligning, gale-church-1993-program} or translation~\cite{moore-2002-fast, varga-2007-parallel,sennrich-volk-2010-mt, gomes-lopes-2016-first}.
The state-of-the art method Vecalign~\cite{thompson-koehn-2019-vecalign} relies on sentence embeddings.

Neither approach is well suited to extracting bitext from grammar books.
Unlike parallel documents, grammar books contain substantial descriptive text surrounding example sentences, making document-level sentence alignment inappropriate.
Bitext mining is more applicable, but relies on multilingual sentence representations that may generalize poorly to unseen and extremely low-resource languages~\cite{artetxe-schwenk-2019-massively,chimoto-bassett-2022-low}

\begin{figure*}[t]
    \centering
    \includegraphics[width=\linewidth]{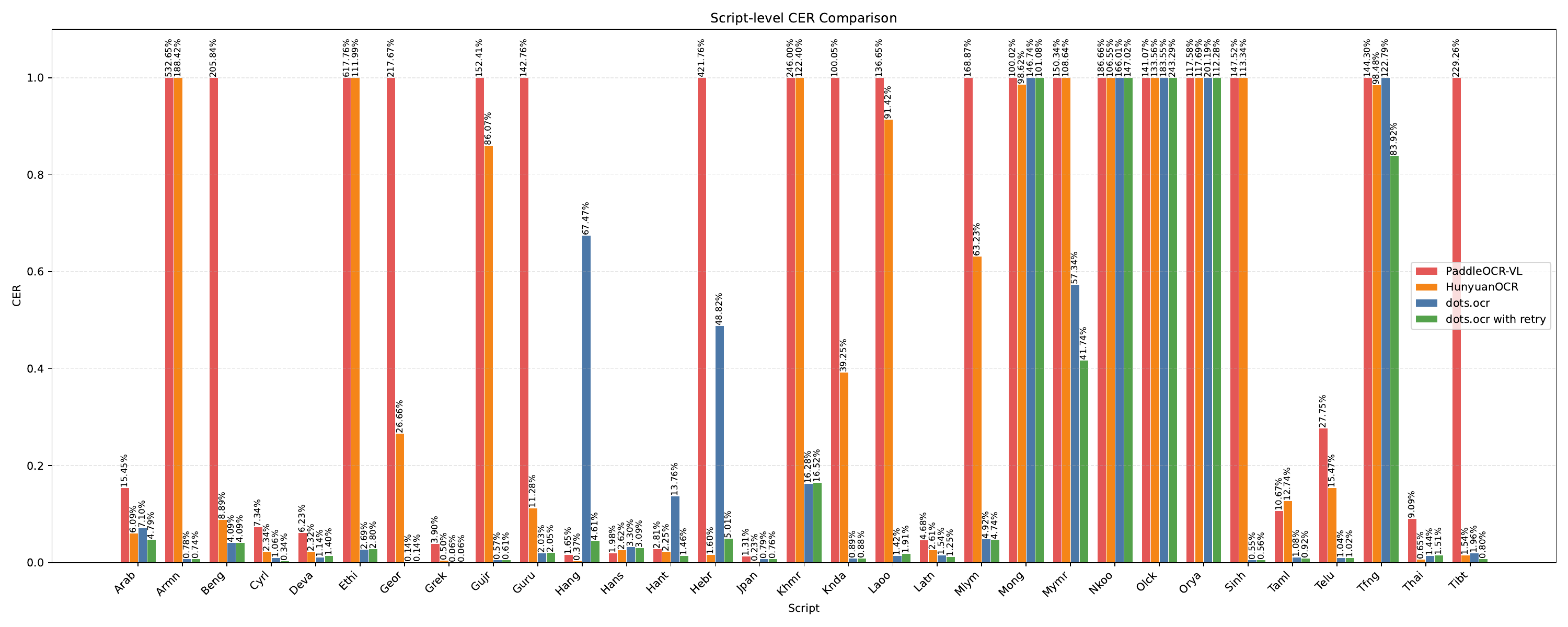}
    \caption{
    Script-level CER for {\dotsocr}, {\hunyuan}, and {\paddle}.
    CER values exceeding 100\% indicate that the corresponding script is not supported by the model.
    }
    \label{fig:script_level_ocr_cer}
\end{figure*}

\section{Script-level OCR Results}
\label{sec:more_ocr}

We report script-level CER in Figure~\ref{fig:script_level_ocr_cer}.
{\dotsocr}, without the retry strategy, already clearly outperforms {\paddle} and {\hunyuan}.
However, we observe that decoding with the default temperature of $0.1$ sometimes degenerate repetition until the maximum generation length is reached.
Therefore, following \citet{poznanski2025olmocr}, we retry decoding with progressively higher temperatures of $0.3$, $0.5$, and $0.8$, whenever the generated sequence reaches the maximum token limit.
This retry strategy substantially improves recognition performance for the \verb|Hang|, \verb|Hant|, and \verb|Hebr| scripts.

\begin{figure}[t]
\small
    \fbox{
        \parbox{0.95\linewidth}{
This document may contain parallel sentences in \texttt{source\_lang} and \texttt{target\_lang}. Extract all paired \texttt{source\_lang}-\texttt{target\_lang} translations. \\

Return a JSON list in the following format: \\
\textasciigrave\textasciigrave\textasciigrave json

[ " \texttt{src\_code}": "<\texttt{source\_lang} sentence>", "\texttt{tgt\_code}": "<\texttt{target\_lang} sentence>", "confidence\_reasoning": "Briefly justify why this is a translation and not an exercise"]

\textasciigrave\textasciigrave\textasciigrave \\

Extraction rules: \\
- Extract a pair only when alignment is visually or structurally indicated (e.g., adjacency, parallel columns, keywords like "translation"). \\
- Understand the content first. If sentences appear only as independent exercises or examples without explicit structural cues linking them, do not extract them. \\
- Sentences appearing in exercises, drills, tests, or prompts (e.g., instructions asking the reader to translate or say the equivalent) are NOT translation pairs and must be ignored. \\
- Do not hallucinate. If more than one plausible alignment exists for a sentence, do not extract it. \\
- If numbering is used (e.g., 1., (2), 3)), numbers must match exactly between \texttt{source\_lang} and \texttt{target\_lang}. If numbering is inconsistent, incomplete, or not strictly one-to-one, do not extract. \\
- Paired sentences must preserve punctuation structure (e.g., ?, !, quotes, colon usage). Mismatched punctuation indicates non-parallel text and the pair should be rejected. \\
- When encountering interlinear glossed text, extract only the original sentence and the translation line, and ignore the morpheme-by-morpheme gloss line. \\
- Each extracted pair must contain exactly one \texttt{source\_lang} sentence and exactly one \texttt{target\_lang} sentence. \\
- Extract only complete sentences.  \\
- Do not translate, paraphrase, or modify sentences. \\
- Enforce strict sequential alignment. Sentence pairs must follow the same order in both languages. Do not skip, reorder, or cross-match sentences. \\
- If no valid parallel sentence pairs are found, return []. \\
- If an extracted pair is determined invalid, include "INVALID" in "confidence\_reasoning". \\

You are given both the document image and its OCR text. Use both sources jointly. If they conflict, prioritize the OCR text. Do not invent or infer missing text.

OCR text: \\
\texttt{OCR\_TEXT}
}
}
    \caption{Instruction template used for bitext extraction.}
    \label{fig:prompt_bitext_extract}
\end{figure}

\begin{figure}[t]
\small
    \fbox{
        \parbox{0.95\linewidth}{
You are a professional translator. Do not say you do not speak the language. Do not say you do not have enough information. You must provide a translation. Do not include explanations, comments, or extra text. The entire response must be a single line containing only the translation. When provided with example translations, use them as reference to help ensure correct translation.
}
}
    \caption{System prompt template used for few-shot experiments.}
    \label{fig:sys_prompt_few_shot}
\end{figure}

\begin{figure}[t]
\small
    \fbox{
        \parbox{0.95\linewidth}{
Translate the following sentence from \texttt{SRC\_LANG} to \texttt{TGT\_LANG}: \texttt{SRC\_SENT}

Below are some example translations that might be useful: \\
\texttt{SRC\_LANG}: \texttt{SRC\_SENT\_1} \\
\texttt{TGT\_LANG}: \texttt{TGT\_SENT\_1} \\

\texttt{SRC\_LANG}: \texttt{SRC\_SENT\_2} \\
\texttt{TGT\_LANG}: \texttt{TGT\_SENT\_2} \\
\ldots \\

Now, please provide the translation. \\
\texttt{SRC\_LANG}: \texttt{TGT\_SENT} \\
\texttt{TGT\_LANG}:
}
}
    \caption{User prompt template used for few-shot experiments.}
    \label{fig:user_prompt_few_shot}
\end{figure}

\section{Prompts}
\label{sec:prompts}

Figure~\ref{fig:prompt_bitext_extract} shows the VLM prompt used for bitext extraction.
Figure~\ref{fig:sys_prompt_few_shot} and Figure~\ref{fig:user_prompt_few_shot} show the system and user prompts used in the few-shot experiments, respectively.

\begin{table*}[t]
\footnotesize
    \centering
    \begin{tabular}{l|p{4.5cm}p{4.5cm}p{4.5cm}}
    \toprule
    \textbf{Lang} & \textbf{Books} & \textbf{Existing Data} & \textbf{Normalization} \\
    \midrule
    \verb|arp| & use diacritics (e.g., é, í, ó) & no diacritics & remove diacritics \\
    \midrule
    \verb|chv| & Latin script & Cyrillic script & script conversion \\
    \midrule
    \verb|ddo| & noisy OCR (e.g., ā, ç, ġ) & diacritics (e.g., ä, č, š) & correct OCR errors \\
    \midrule
    \verb|dgo| & IPA & Devanagari script & Codex-proposed conversion rules \\
    \midrule
    \verb|dik| & diacritics and noisy OCR (e.g., à, ç, ē) & diacritics (e.g., í, ó, ú) & correct OCR errors \\
    \midrule
    \verb|git| & use diacritics & no diacritics; use U+0332 & remove diacritics \\
    \midrule
    \verb|gug| & diacritics (e.g., ç, ë, ô) & diacritics (e.g., ö, û, ü) & correct OCR errors \\
    \midrule
    \verb|ilo| & diacritics (e.g., ã, ç, í) & few diacritics & remove diacritics \\
    \midrule
    \verb|kac| & diacritics (e.g., ă, ė, ū) & few diacritics & remove diacritics \\
    \midrule
    \verb|kea| & \multicolumn{2}{c}{diacritics largely consistent} & none \\
    \midrule
    \verb|kgv| & \multicolumn{2}{c}{orthographies consistent} & none \\
    \midrule
    \verb|kmb| & diacritics (e.g., ā, ġ, ň) & fewer diacritics (e.g., ã, ñ, ó) & correct OCR errors \\
    \midrule
    \verb|lez| & Latin script & Cyrillic script & script conversion \\
    \midrule
    \verb|lus| & diacritics (e.g., á, ì, ě) & fewer diacritics (e.g., á, é, î) & normalize diacritics \\
    \midrule
    \verb|mgm| & diacritics (e.g., í, ó, ú) & few diacritics & remove diacritics \\
    \midrule
    \verb|min| & \multicolumn{2}{c}{orthographies consistent} & none \\
    \midrule
    \verb|mnc| & diacritics (e.g., ī, š, ū) & no diacritics & remove diacritics from books  \\
    \midrule
    \verb|ntu| & diacritics (e.g., ä, ë, ö) & no diacritics & normalize diacritics \\
    \midrule
    \verb|usp| & \multicolumn{2}{c}{diacritics largely consistent} & none \\
    \midrule
    \verb|wol| & diacritics (e.g., å, ņ, ū) & diacritics (e.g., î, ô, ü) & normalize diacritics \\
    \bottomrule
    \end{tabular}
    \caption{
    Orthographic differences between the archival books and the existing dataset, along with the normalization applied to reconcile them. Differences caused by OCR errors are also included.
    }
    \label{tab:lang_ortho_norm}
\end{table*}

\section{Orthography Normalization for Main Results}
\label{sec:ortho_norm_detail}
We summarize the orthography normalization strategies in Table~\ref{tab:lang_ortho_norm}.
The most salient challenge is the inconsistent usage of diacritics between the archival books and existing datasets.
We manually reconcile some of these inconsistencies.
However, the normalization is not exhaustive, and mismatches in diacritic usage still remain.

\begin{table}[t]
\footnotesize
    \centering
    \begin{tabular}{l|rr}
    \toprule
    \textbf{Lang} & \textbf{Standard Norm.} & \textbf{Further Norm.} \\
    \midrule
    \multicolumn{3}{l}{\textit{Into-LRL translation}} \\
    \verb|ddo| & 24.24 & \upcell{25.86}{1.62} \\
    \verb|dik| & 16.89 & \upcell{19.92}{3.03} \\
    \verb|git| & 23.16 & \upcell{25.14}{1.98} \\
    \verb|gug| & 26.05 & \downcell{24.63}{1.42} \\
    \verb|kmb| & 16.19 & \upcell{16.52}{0.33} \\
    \verb|wol| & 26.93 & \downcell{26.63}{0.30} \\
    \midrule
    \multicolumn{3}{l}{\textit{Into-HRL translation}} \\
    \verb|ddo| & 27.64 & \downcell{26.99}{0.65} \\
    \verb|dik| & 25.39 & \upcell{25.66}{0.27} \\
    \verb|git| & 30.49 & \downcell{29.09}{1.40} \\
    \verb|gug| & 44.06 & \downcell{42.32}{1.74} \\
    \verb|kmb| & 25.42 & \downcell{24.40}{1.02} \\
    \verb|wol| & 39.75 & \downcell{38.31}{1.44} \\
    \bottomrule
    \end{tabular}
    \caption{
    512 shot chrF2++ scores with \texttt{Qwen}.
    Standard Norm. applies only the normalization steps in Section~\ref{sec:post_align} to the extracted bitext, while Further Norm. additionally removes diacritics from both the extracted and test sentences.
    }
    \label{tab:furthur_norm}
\end{table}

\subsection{Normalizing Special Letters for both Example and Test Sentences}
\label{sec:further_norm}

Since OCR systems often fail to recognize diacritics and special letters, such as the Combining Low Line character (U+0332), we examine whether these OCR errors substantially affect downstream MT performance.
Specifically, we remove diacritics and the Combining Low Line character from both the training and test sets and re-evaluate our few-shot MT experiments for the affected languages: \verb|ddo|, \verb|dik|, \verb|git|, \verb|gug|, \verb|kmb|, \verb|wol|.
Our motivation is text will remain intelligible even when diacritics are omitted.

Table~\ref{tab:furthur_norm} compares performance before and after this additional normalization.
The results are very similar across the two settings, suggesting that errors involving these characters have only a limited impact on MT performance in our experiments.

\section{Applying Existing Alignment Methods}
\label{sec:alignment_details}

\paragraph{Preprocessing}
\label{par:langid}
Sentence alignment methods assume two documents containing the same content in different languages.
Linguistic resources such as grammar books violate this assumption, so we split the OCR text into two language-specific documents.
We segment the text into sentences using a sentence splitter,\footnote{\url{https://github.com/mediacloud/sentence-splitter}} and then apply the langID model from~\citet{nllbteam2022languageleftbehindscaling} to each sentence.
Because the langID model does not support all low-resource languages, we only trust predictions of high-resource languages.
For \verb|mgm-eng|, for example, sentences classified as \verb|eng| are assigned to English, while all other sentences are assigned to \verb|mgm|.
We then split the sentences into separate files for each language.

\paragraph{Sentence Mining}
We consider two variants of sentence mining: \textit{Global Mining}, applied separately to each book, and \textit{Local Mining}, applied separately to each page.
For each pair of source and target sentences, we compute the margin score (Eq.~\ref{eq:margin_score}) and keep the highest-scoring pairs.
Our implementation is based on \verb|stopes|~\cite{andrews-etal-2022-stopes}.

\paragraph{Vecalign}
We apply Vecalign separately to each page, using the original implementation.\footnote{\url{https://github.com/thompsonb/vecalign}}
For simplicity, we disable many-to-many alignments and  allow only 1-1 alignments.

\paragraph{Hunalign}
Due to the poor generalization of sentence embeddings, we also consider non-embedding-based Hunalign~\cite{varga-2007-parallel}, previously used in \citet{merx-etal-2024-low} for Mambai alignment.
Hunalign relies on a dictionary and sentence lengths.
We reuse the dictionary entries from \citet{merx-etal-2024-low}, which were extract based on text formatting.
But we note that dictionaries are often not readily available in practice.
We apply Hunalign separately to each page.

For Sentence Mining and Vecalign, we tune the score thresholds to maximize F1.

\section{Additional Results}

\begin{table}[t]
\small
    \centering
    \begin{tabular}{l|rr}
    \toprule
    \multirow{2}{*}{\textbf{Lang}} & \multicolumn{2}{c}{\textbf{BLASER 2.0-QE$\uparrow$}} \\
    \cmidrule{2-3}
    & \textbf{Extracted} & \textbf{Test} \\
    \midrule
    \verb|arp| & 2.798 & 2.755 \\
    \verb|chv| & 2.804 & 2.792 \\
    \verb|ddo| & 2.721 & 2.466 \\
    \verb|dgo| & 2.851 & 3.683 \\
    \verb|git| & 2.867 & 2.820 \\
    \verb|kgv| & 2.667 & 2.769 \\
    \verb|lez| & 2.702 & 2.556 \\
    \verb|mgm| & 2.941 & 3.048 \\
    \verb|mnc| & 2.558 & 2.514 \\
    \verb|ntu| & 2.406 & 2.468 \\
    \verb|usp| & 2.148 & 2.458 \\
    \midrule
    Avg. & 2.678 & 2.757 \\
    \bottomrule
    \end{tabular}
    \caption{BLASER 2.0-QE scores for unsupported languages.}
    \label{tab:qe_scores_unsupported}
\end{table}

\subsection{Intrinsic Evaluation for Unseen Languages}
\label{sec:blaser_unsupported}

We report BLASER 2.0-QE scores for unsupported languages in Table~\ref{tab:qe_scores_unsupported}.
The quality gap between the extracted examples and the test examples is relatively small.
However, because BLASER 2.0-QE may itself be unreliable for these unsupported languages, we treat these scores only as a rough reference and do not use them as a primary evaluation metric.

\begin{table}[t]
\footnotesize
    \centering
    \begin{tabular}{l|rr|rr}
    \toprule
    \multirow{2}{*}{\textbf{Lang}} & \multicolumn{2}{c|}{\textbf{Into LRL}} & \multicolumn{2}{c}{\textbf{Into HRL}} \\
    \cmidrule(lr){2-3} \cmidrule(lr){4-5}
    & \multicolumn{1}{c}{0 shot} & \multicolumn{1}{c}{8 shot} & \multicolumn{1}{c}{0 shot} & \multicolumn{1}{c}{8 shot} \\
    \midrule
\multicolumn{5}{l}{\textit{Without NLLB support}} \\
\verb|arp| & 5.02 & \textsuperscript{*}\upcell{23.39}{18.37} & 11.88 & \textsuperscript{*}\upcell{16.72}{4.84} \\
\verb|chv| & 15.56 & \textsuperscript{*}\upcell{18.29}{2.73} & 33.42 & \textsuperscript{*}\upcell{33.94}{0.52} \\
\verb|ddo| & 0.60 & \textsuperscript{*}\upcell{13.46}{12.86} & 16.19 & \textsuperscript{*}\upcell{19.56}{3.37} \\
\verb|dgo| & 13.08 & \textsuperscript{*}\upcell{16.59}{3.51} & 36.84 & \downcell{36.60}{0.24} \\
\verb|git| & 7.18 & \textsuperscript{*}\upcell{11.47}{4.29} & 23.96 & \upcell{25.57}{1.61} \\
\verb|kgv| & 12.46 & \textsuperscript{*}\upcell{25.53}{13.07} & 14.36 & \textsuperscript{*}\upcell{22.76}{8.40} \\
\verb|lez| & 4.33 & \textsuperscript{*}\upcell{16.08}{11.75} & 18.87 & \upcell{19.82}{0.95} \\
\verb|mgm| & 12.51 & \textsuperscript{*}\upcell{23.85}{11.34} & 26.43 & \textsuperscript{*}\upcell{30.19}{3.76} \\
\verb|mnc| & 0.00 & \textsuperscript{*}\upcell{17.17}{17.17} & 12.81 & \textsuperscript{*}\upcell{15.88}{3.07} \\
\verb|ntu| & 7.31 & \textsuperscript{*}\upcell{30.53}{23.22} & 15.26 & \textsuperscript{*}\upcell{28.17}{12.91} \\
\verb|usp| & 9.57 & \textsuperscript{*}\upcell{17.49}{7.92} & 11.73 & \textsuperscript{*}\upcell{17.70}{5.97} \\
\midrule
\multicolumn{5}{l}{\textit{With NLLB support}} \\
\verb|dik| & 9.98 & \textsuperscript{*}\upcell{11.75}{1.77} & 22.19 & \textsuperscript{*}\downcell{21.66}{0.53} \\
\verb|gug| & 16.56 & \textsuperscript{*}\upcell{18.17}{1.61} & 30.12 & \textsuperscript{*}\upcell{30.61}{0.49} \\
\verb|ilo| & 33.26 & \textsuperscript{*}\upcell{34.64}{1.38} & 44.25 & \downcell{44.14}{0.11} \\
\verb|kac| & 6.07 & \textsuperscript{*}\upcell{18.87}{12.80} & 22.30 & \upcell{22.50}{0.20} \\
\verb|kea| & 24.92 & \textsuperscript{*}\upcell{28.29}{3.37} & 51.52 & \textsuperscript{*}\upcell{53.13}{1.61} \\
\verb|kmb| & 13.09 & \textsuperscript{*}\downcell{12.21}{0.88} & 20.20 & \upcell{20.48}{0.28} \\
\verb|lus| & 25.56 & \textsuperscript{*}\downcell{22.00}{3.56} & 31.43 & \textsuperscript{*}\downcell{30.94}{0.49} \\
\verb|min| & 34.78 & \upcell{34.83}{0.05} & 48.30 & \upcell{48.49}{0.19} \\
\verb|wol| & 14.08 & \textsuperscript{*}\upcell{15.37}{1.29} & 24.88 & \downcell{24.81}{0.07} \\
    \midrule
    Avg. & 13.30 & \upcell{20.50}{7.20} & 25.85 & \upcell{28.18}{2.33} \\
    \bottomrule
    \end{tabular}
    \caption{chrF2++ ($\uparrow$) scores for few-shot MT using \texttt{Llama-3.1-8B}.
    \textsuperscript{*}$\text{p-value} <0.05$.
    }
    \label{tab:few_shot_mt_llama}
\end{table}

\subsection{Few-shot Experiments using Llama}
\label{sec:few_shot_llama}

We provide additional few-shot results on the 20 main languages using a Llama model (\verb|meta-llama/Llama-3.1-8B|~\cite{grattafiori2024llama3herdmodels}).

Table~\ref{tab:few_shot_mt_llama} reports the results.
We observe improvements for most translation directions, with average gains of 7.20 and 2.33 chrF2++ for translation in LRL and HRL, respectively.
These results suggest that the extracted examples provide substantial benefits even when used with a considerably smaller, less powerful model.

\begin{figure}
    \centering
    \includegraphics[width=\linewidth]{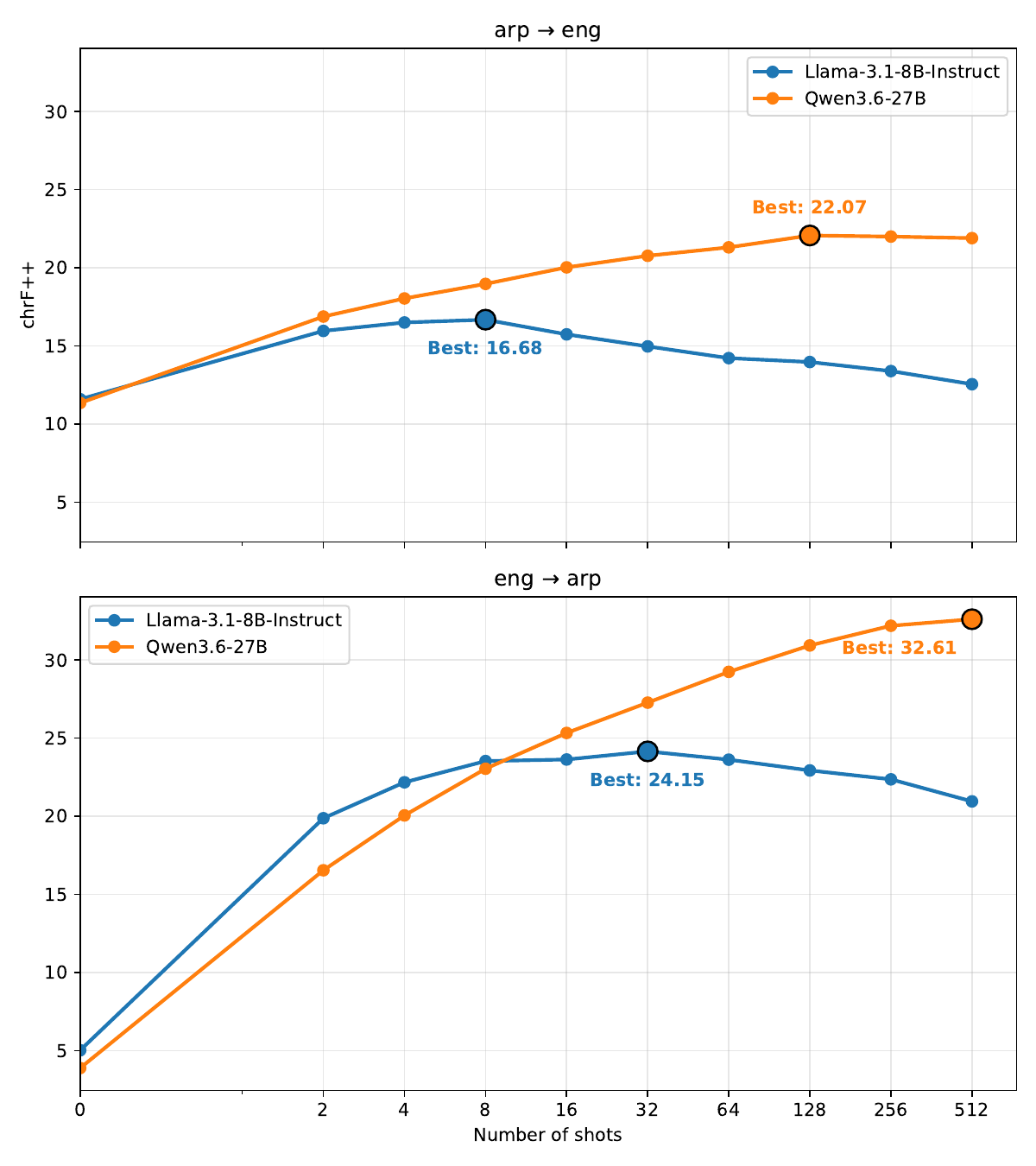}
    \caption{
    Few-shot MT experiments on \texttt{arp} dev set with varying numbers of parallel examples.
    Performance of \texttt{Qwen3.6-27B} continues to improve as more examples are provided, whereas \texttt{Llama-3.1-8B-instruct} plateaus quickly.
    }
    \label{fig:few_shot_ablation}
\end{figure}

\subsection{Optimize the Number of Bitext Examples}
\label{sec:few_shot_ablation}

We vary the number of bitext examples from 2 to up to 512 to examine how few-shot MT performance changes with the number of examples.
We evaluate \texttt{Llama-3.1-8B} and \texttt{Qwen3.6-27B} on the \texttt{arp-eng} dev set.

As shown in Figure~\ref{fig:few_shot_ablation}, performance of \texttt{Llama-3.1-8B} saturates quickly and plateaus at approximately 8 examples, whereas performance of \texttt{Qwen3.6-27B} continues to improve.
Therefore, we perform 512-shot experiment for \texttt{Qwen3.6-27B}  and 8-shot for \texttt{Llama-3.1-8B}.
We do not evaluate more than 512 examples because gains diminish while memory demands and inference time become prohibitive.

\begin{table}[t]
\footnotesize
    \centering
    \begin{tabular}{l|rr|rr}
    \toprule
    \multirow{2}{*}{\textbf{Lang}} & \multicolumn{2}{c|}{\texttt{Qwen}} & \multicolumn{2}{c}{\texttt{nllb}} \\
    \cmidrule(lr){2-3} \cmidrule(lr){4-5} 
    & \multicolumn{1}{c}{0 shot} & \multicolumn{1}{c}{512 shot} & \multicolumn{1}{c}{OOB} & \multicolumn{1}{c}{fine-tuned} \\
    \midrule
\multicolumn{5}{l}{\textit{Without NLLB support}} \\
\verb|arp| & 0.14 & \textsuperscript{*}\upcell{16.73}{16.59} & -- & 1.00 \\
\verb|chv| & 6.69 & \textsuperscript{*}\upcell{20.85}{14.16} & -- & 0.42 \\
\verb|ddo| & 0.00 & \textsuperscript{*}\upcell{11.83}{11.83} & -- & 0.08 \\
\verb|dgo| & 4.21 & \downcell{4.10}{0.11} & -- & 0.24 \\
\verb|git| & 0.07 & \textsuperscript{*}\upcell{6.48}{6.41} & -- & 0.08 \\
\verb|kgv| & 0.23 & \textsuperscript{*}\upcell{18.59}{18.36} & -- & 0.66 \\
\verb|lez| & 0.74 & \textsuperscript{*}\upcell{14.92}{14.18} & -- & 0.17 \\
\verb|mgm| & 2.16 & \textsuperscript{*}\upcell{10.81}{8.65} & -- & 1.98 \\
\verb|mnc| & 0.17 & \textsuperscript{*}\upcell{7.37}{7.20} & -- & 0.04 \\
\verb|ntu| & 0.36 & \textsuperscript{*}\upcell{34.69}{34.33} & -- & 1.10 \\
\verb|usp| & 0.07 & \textsuperscript{*}\upcell{10.28}{10.21} & -- & 0.31 \\
\textit{Avg}. & \textit{1.35} & \upcell{\textit{14.24}}{12.89} & -- & \textit{0.55} \\
\midrule
\multicolumn{5}{l}{\textit{With NLLB support}} \\
\verb|dik| & 0.70 & \textsuperscript{*}\upcell{4.40}{3.70} & 3.18 & \textsuperscript{*}\downcell{1.66}{1.52} \\
\verb|gug| & 1.83 & \textsuperscript{*}\upcell{6.66}{4.83} & 17.12 & \textsuperscript{*}\downcell{13.31}{3.81} \\
\verb|ilo| & 15.19 & \textsuperscript{*}\upcell{19.44}{4.25} & 27.03 & \textsuperscript{*}\downcell{9.39}{17.64} \\
\verb|kac| & 0.76 & \textsuperscript{*}\upcell{6.72}{5.96} & 14.92 & \textsuperscript{*}\downcell{7.64}{7.28} \\
\verb|kea| & 5.64 & \textsuperscript{*}\upcell{13.64}{8.00} & 23.69 & \textsuperscript{*}\downcell{18.09}{5.60} \\
\verb|kmb| & 1.71 & \downcell{1.60}{0.11} & 4.31 & \textsuperscript{*}\downcell{0.53}{3.78} \\
\verb|lus| & 1.60 & \textsuperscript{*}\upcell{5.23}{3.63} & 13.28 & \textsuperscript{*}\downcell{0.85}{12.43} \\
\verb|min| & 12.15 & \textsuperscript{*}\upcell{16.50}{4.35} & 24.05 & \textsuperscript{*}\downcell{8.66}{15.39} \\
\verb|wol| & 4.72 & \textsuperscript{*}\upcell{7.64}{2.92} & 6.53 & \textsuperscript{*}\downcell{1.60}{4.93} \\
\textit{Avg}. & \textit{4.92} & \upcell{\textit{9.09}}{4.17} & \textit{14.90} & \downcell{\textit{6.86}}{8.04} \\
    \midrule
    \textbf{Avg}. & 2.96 & \upcell{11.92}{8.96} & -- & 3.39  \\
    \bottomrule
    \end{tabular}
    \caption{spBLEU scores for translation \textit{into LRL}.
    \textsuperscript{*}$\text{p-value} <0.05$.
    }
    \label{tab:into_lrl_mt_spbleu}
\end{table}

\begin{table}[t]
\footnotesize
    \centering
    \begin{tabular}{l|rr|rr}
    \toprule
    \multirow{2}{*}{\textbf{Lang}} & \multicolumn{2}{c|}{\texttt{Qwen}} & \multicolumn{2}{c}{\texttt{nllb}} \\
    \cmidrule(lr){2-3} \cmidrule(lr){4-5} 
    & \multicolumn{1}{c}{0 shot} & \multicolumn{1}{c}{512 shot} & \multicolumn{1}{c}{OOB} & \multicolumn{1}{c}{fine-tuned} \\
    \midrule
\multicolumn{5}{l}{\textit{Without NLLB support}$^\dagger$} \\
\verb|arp| & 2.37 & \textsuperscript{*}\upcell{7.98}{5.61} & 0.67 & \textsuperscript{*}\upcell{1.57}{0.90} \\
\verb|chv| & 25.00 & \textsuperscript{*}\upcell{26.23}{1.23} & 2.76 & \textsuperscript{*}\upcell{5.10}{2.34} \\
\verb|ddo| & 0.31 & \textsuperscript{*}\upcell{8.35}{8.04} & 0.20 & \textsuperscript{*}\upcell{1.47}{1.27} \\
\verb|dgo| & 24.53 & \textsuperscript{*}\upcell{27.76}{3.23} & 19.39 & \textsuperscript{*}\downcell{14.09}{5.30} \\
\verb|git| & 7.78 & \textsuperscript{*}\upcell{10.87}{3.09} & 1.40 & \textsuperscript{*}\upcell{6.17}{4.77} \\
\verb|kgv| & 3.54 & \textsuperscript{*}\upcell{20.58}{17.04} & 1.27 & \textsuperscript{*}\upcell{7.32}{6.05} \\
\verb|lez| & 6.41 & \textsuperscript{*}\upcell{10.97}{4.56} & 0.50 & \textsuperscript{*}\upcell{2.55}{2.05} \\
\verb|mgm| & 11.00 & \textsuperscript{*}\upcell{24.78}{13.78} & 3.66 & \textsuperscript{*}\upcell{13.29}{9.63} \\
\verb|mnc| & 2.38 & \textsuperscript{*}\upcell{5.43}{3.05} & 0.10 & \upcell{0.56}{0.46} \\
\verb|ntu| & 2.92 & \textsuperscript{*}\upcell{21.42}{18.50} & 0.58 & \textsuperscript{*}\upcell{18.25}{17.67} \\
\verb|usp| & 0.67 & \textsuperscript{*}\upcell{7.05}{6.38} & 0.34 & \textsuperscript{*}\upcell{2.21}{1.87} \\
\textit{Avg}. & \textit{7.90} & \upcell{\textit{15.58}}{7.68} & \textit{2.81} & \upcell{\textit{6.60}}{3.79} \\
\midrule
\multicolumn{5}{l}{\textit{With NLLB support}} \\
\verb|dik| & 9.50 & \textsuperscript{*}\upcell{10.61}{1.11} & 14.14 & \textsuperscript{*}\downcell{11.70}{2.44} \\
\verb|gug| & 23.13 & \textsuperscript{*}\upcell{24.83}{1.70} & 28.41 & \textsuperscript{*}\downcell{24.17}{4.24} \\
\verb|ilo| & 36.60 & \textsuperscript{*}\upcell{38.64}{2.04} & 41.72 & \textsuperscript{*}\downcell{38.66}{3.06} \\
\verb|kac| & 7.88 & \textsuperscript{*}\upcell{8.51}{0.63} & 18.26 & \textsuperscript{*}\downcell{15.27}{2.99} \\
\verb|kea| & 46.58 & \textsuperscript{*}\upcell{51.98}{5.40} & 50.41 & \textsuperscript{*}\downcell{48.20}{2.21} \\
\verb|kmb| & 9.10 & \upcell{9.33}{0.23} & 13.78 & \textsuperscript{*}\downcell{11.06}{2.72} \\
\verb|lus| & 16.00 & \textsuperscript{*}\upcell{17.01}{1.01} & 22.29 & \textsuperscript{*}\downcell{18.75}{3.54} \\
\verb|min| & 38.20 & \textsuperscript{*}\upcell{41.07}{2.87} & 41.17 & \textsuperscript{*}\downcell{35.77}{5.40} \\
\verb|wol| & 18.79 & \textsuperscript{*}\upcell{20.56}{1.77} & 20.92 & \textsuperscript{*}\downcell{16.44}{4.48} \\
\textit{Avg}. & \textit{22.86} & \upcell{\textit{24.73}}{1.87} & \textit{27.90} & \downcell{\textit{24.45}}{3.45} \\
    \midrule
    Avg. & 14.63 & \upcell{19.70}{5.07} & 14.10 & \upcell{14.63}{0.53} \\
    \bottomrule
    \end{tabular}
    \caption{
    spBLEU scores for translation \textit{into HRL}.
    $^\dagger$Proxy source language codes are used.
    \textsuperscript{*}$\text{p-value} <0.05$.
    }
    \label{tab:into_hrl_mt_spbleu}
\end{table}

\subsection{spBLEU Scores for the Main Results}

We report spBLEU scores for translation into LRLs and HRLs in Table~\ref{tab:into_lrl_mt_spbleu} and Table~\ref{tab:into_hrl_mt_spbleu}, respectively, for the 20 main languages.

The conclusions are similar to those in the main results.
Few-shot MT yields substantial gains compared to zero-shot, with average improvements of 8.96 and 5.07 spBLEU for translation into LRLs and HRLs, respectively.
Fine-tuning \texttt{nllb} hurts its translation performance on languages it originally supports, but improves translation into HRLs for newly supported languages.

\begin{table*}[t]
\small
\includegraphics[width=0.8\textwidth]{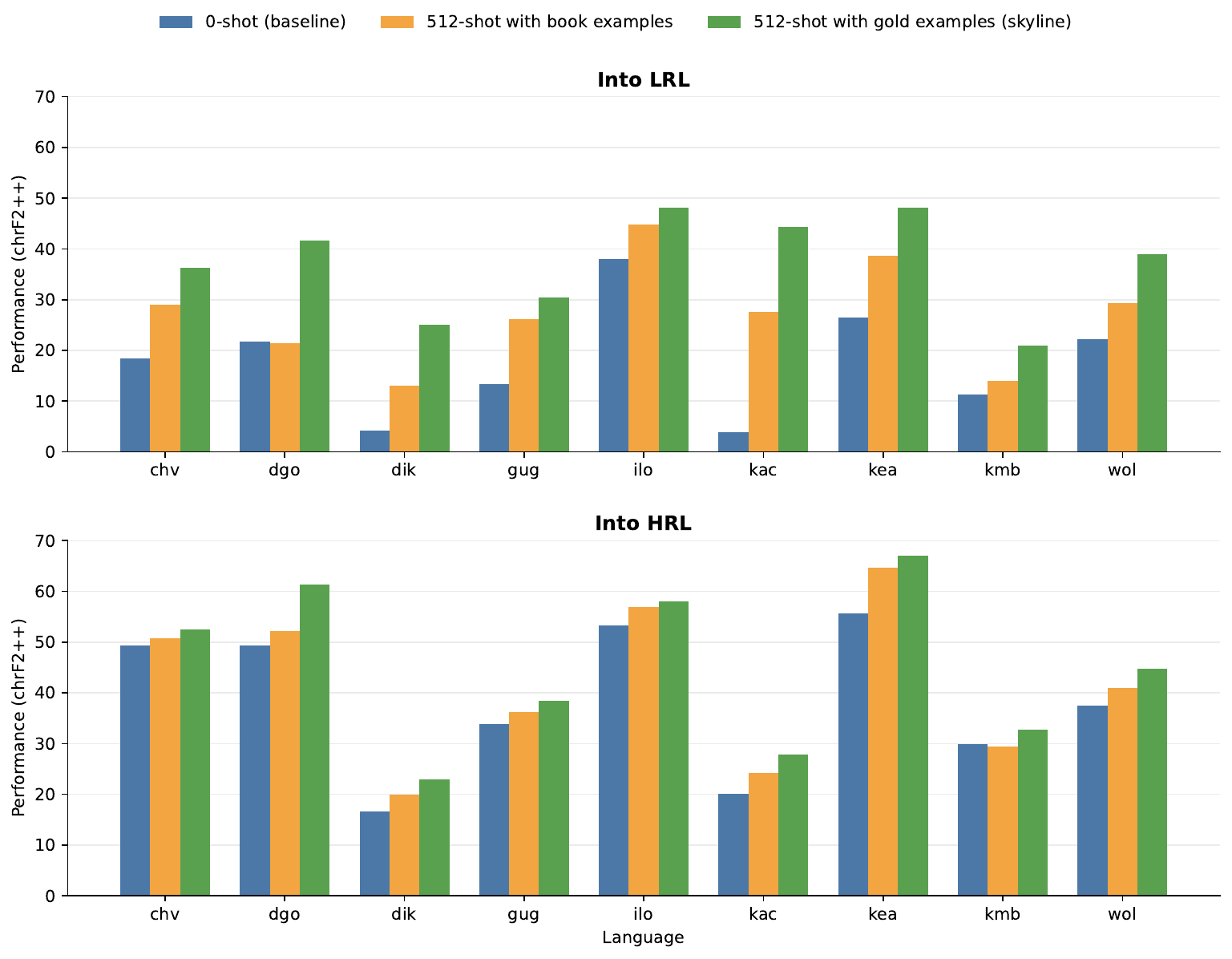}
\vspace{0.5cm}
    \centering
    \begin{tabular}{l|rrr|rrr}
    \toprule
    \multirow{3}{*}{\textbf{Lang}} & \multicolumn{3}{c}{\textbf{Into LRL}} & \multicolumn{3}{c}{\textbf{Into HRL}} \\
    \cmidrule(lr){2-4} \cmidrule(lr){5-7}
    & \multirow{2}{*}{0 shot} & \multicolumn{2}{c}{512 shot} & \multirow{2}{*}{0 shot} & \multicolumn{2}{c}{512 shot} \\
    \cmidrule(lr){3-4} \cmidrule(lr){6-7}
    & & \multicolumn{1}{c}{Book} & \multicolumn{1}{c}{Gold} & & \multicolumn{1}{c}{Book} & \multicolumn{1}{c}{Gold} \\
    \cmidrule{1-1} \cmidrule(lr){2-4} \cmidrule(lr){5-7}
\multicolumn{7}{l}{\textit{Without NLLB support}} \\
\verb|chv| & 18.43 & \textsuperscript{*}\upcell{29.03}{10.60} & \textsuperscript{*}\upcell{36.30}{7.27} & 49.41 & \textsuperscript{*}\upcell{50.76}{1.35} & \textsuperscript{*}\upcell{52.45}{1.69} \\
\verb|dgo| & 21.79 & \textsuperscript{*}\downcell{21.44}{0.35} & \textsuperscript{*}\upcell{41.62}{20.18} & 49.34 & \textsuperscript{*}\upcell{52.11}{2.77} & \textsuperscript{*}\upcell{61.37}{9.26} \\
\midrule
\multicolumn{7}{l}{\textit{With NLLB support}} \\
\verb|dik| & 4.15 & \textsuperscript{*}\upcell{13.07}{8.92} & \textsuperscript{*}\upcell{25.10}{12.03} & 16.68 & \textsuperscript{*}\upcell{19.98}{3.30} & \textsuperscript{*}\upcell{23.00}{3.02} \\
\verb|gug| & 13.43 & \textsuperscript{*}\upcell{26.10}{12.67} & \textsuperscript{*}\upcell{30.36}{4.26} & 33.90 & \textsuperscript{*}\upcell{36.28}{2.38} & \textsuperscript{*}\upcell{38.41}{2.13} \\
\verb|ilo| & 38.05 & \textsuperscript{*}\upcell{44.77}{6.72} & \textsuperscript{*}\upcell{48.16}{3.39} & 53.27 & \textsuperscript{*}\upcell{57.00}{3.73} & \textsuperscript{*}\upcell{57.98}{0.98} \\
\verb|kac| & 3.93 & \textsuperscript{*}\upcell{27.50}{23.57} & \textsuperscript{*}\upcell{44.33}{16.83} & 20.14 & \textsuperscript{*}\upcell{24.27}{4.13} & \textsuperscript{*}\upcell{27.87}{3.60} \\
\verb|kea| & 26.50 & \textsuperscript{*}\upcell{38.61}{12.11} & \textsuperscript{*}\upcell{48.12}{9.51} & 55.68 & \textsuperscript{*}\upcell{64.61}{8.93} & \textsuperscript{*}\upcell{66.97}{2.36} \\
\verb|kmb| & 11.34 & \textsuperscript{*}\upcell{13.96}{2.62} & \textsuperscript{*}\upcell{20.87}{6.91}  & 29.83 & \downcell{29.41}{0.42} & \textsuperscript{*}\upcell{32.79}{3.38} \\
\verb|wol| & 22.15 & \textsuperscript{*}\upcell{29.28}{7.13} & \textsuperscript{*}\upcell{39.03}{9.75} & 37.48 & \textsuperscript{*}\upcell{41.03}{3.55} & \textsuperscript{*}\upcell{44.74}{3.71} \\
\midrule
Avg. & 17.75 & \upcell{27.08}{9.33} & \upcell{37.10}{10.02} & 38.41 & \upcell{41.72}{3.31} & \upcell{45.06}{3.34} \\
\bottomrule
    \end{tabular}
    \caption{
    chrF2++ ($\uparrow$) results for \texttt{Qwen3.6-27B} on the BOUQuET test set.
    0-shot performance serves as the baseline for comparison with 512-shot using book examples.
    The 512-shot setting with gold in-domain examples serves as the topline for comparison with 512-shot prompting using book examples.
    \textsuperscript{*}$\text{p-value} <0.05$.
    On the top is a visualization of the results.
    }
    \label{tab:chrf_bouquet}
\end{table*}

\subsection{Evaluation Results on BOUQuET}
\label{sec:bouquet_result}

We compare 512-shot translation using Qwen3.6-27B against its 0-shot performance on the BOUQuET test set.
For the topline setting, we use in-context examples drawn from the BOUQuET dev set.

Table~\ref{tab:chrf_bouquet} reports chrF2++ scores for translation into LRLs and HRLs.
Using parallel sentences from books as in-context examples yields substantial improvements, with average gains of 9.33 and 3.31 chrF2++ for translation into LRLs and HRLs, respectively.

However, a substantial gap remains relative to the topline setting, which is 10.02 chrF2++ for translation into LRLs and 3.34 chrF2++ for translation into HRLs.

\begin{table}[t]
\small
    \centering
    \begin{tabular}{l|rr|rr}
    \toprule
    \multirow{2}{*}{\textbf{Lang}} & \multicolumn{2}{c}{\textbf{Into Creole}} &  \multicolumn{2}{c}{\textbf{Into English}} \\ 
    & \multicolumn{1}{c}{0 shot} & \multicolumn{1}{c}{512 shot} & \multicolumn{1}{c}{0 shot} & \multicolumn{1}{c}{512 shot} \\
    \cmidrule{1-1} \cmidrule(lr){2-3} \cmidrule(lr){4-5} 
\verb|aoa| & 9.64 & \textsuperscript{*}\upcell{23.35}{13.71} & 10.90 & \textsuperscript{*}\upcell{23.62}{12.72} \\
\verb|bah| & 11.77 & \textsuperscript{*}\downcell{7.61}{4.16} & 51.31 & \upcell{51.89}{0.58} \\
\verb|brc| & 10.52 & \textsuperscript{*}\upcell{39.29}{28.77} & 15.23 & \textsuperscript{*}\upcell{44.71}{29.48} \\
\verb|bzj| & 25.12 & \textsuperscript{*}\upcell{28.25}{3.13} & 38.73 & \textsuperscript{*}\upcell{48.14}{9.41} \\
\verb|bzk| & 13.10 & \textsuperscript{*}\upcell{36.45}{23.35} & 57.61 & \textsuperscript{*}\upcell{62.41}{4.80} \\
\verb|cri| & 11.05 & \textsuperscript{*}\upcell{23.27}{12.22} & 9.68 & \textsuperscript{*}\upcell{20.96}{11.28} \\
\verb|dcr| & 10.91 & \textsuperscript{*}\upcell{34.42}{23.51} & 28.93 & \textsuperscript{*}\upcell{56.06}{27.13} \\
\verb|fab| & 9.51 & \textsuperscript{*}\upcell{22.82}{13.31} & 10.07 & \textsuperscript{*}\upcell{23.67}{13.60} \\
\verb|fng| & 12.65 & \textsuperscript{*}\upcell{52.28}{39.63} & 17.76 & \textsuperscript{*}\upcell{31.73}{13.97} \\
\verb|fpe| & 17.16 & \textsuperscript{*}\upcell{31.82}{14.66} & 27.71 & \textsuperscript{*}\upcell{57.80}{30.09} \\
\verb|gcr| & 22.86 & \textsuperscript{*}\upcell{48.71}{25.85} & 46.71 & \textsuperscript{*}\upcell{54.16}{7.45} \\
\verb|gpe| & 13.83 & \textsuperscript{*}\upcell{16.02}{2.19} & 40.62 & \textsuperscript{*}\upcell{55.44}{14.82} \\
\verb|gul| & 38.44 & \textsuperscript{*}\downcell{32.74}{5.70} & 67.30 & \upcell{68.10}{0.80} \\
\verb|gyn| & 15.73 & \textsuperscript{*}\upcell{41.12}{25.39} & 34.94 & \textsuperscript{*}\upcell{45.73}{10.79} \\
\verb|icr| & 9.90 & \textsuperscript{*}\upcell{53.45}{43.55} & 30.10 & \textsuperscript{*}\upcell{64.71}{34.61} \\
\verb|jam| & 25.09 & \textsuperscript{*}\upcell{41.20}{16.11} & 84.90 & \textsuperscript{*}\upcell{88.82}{3.92} \\
\verb|kri| & 21.62 & \textsuperscript{*}\upcell{45.67}{24.05} & 38.64 & \textsuperscript{*}\upcell{47.39}{8.75} \\
\verb|ktu| & 13.11 & \textsuperscript{*}\upcell{50.84}{37.73} & 27.78 & \textsuperscript{*}\upcell{43.98}{16.20} \\
\verb|mue| & 10.26 & \textsuperscript{*}\upcell{18.55}{8.29} & 16.40 & \textsuperscript{*}\upcell{24.56}{8.16} \\
\verb|pov| & 18.12 & \textsuperscript{*}\upcell{35.52}{17.40} & 33.06 & \textsuperscript{*}\upcell{49.19}{16.13} \\
\verb|pre| & 10.95 & \textsuperscript{*}\upcell{51.38}{40.43} & 14.40 & \textsuperscript{*}\upcell{57.31}{42.91} \\
\verb|rcf| & 19.10 & \textsuperscript{*}\upcell{23.99}{4.89} & 41.46 & \textsuperscript{*}\upcell{46.42}{4.96} \\
\verb|srm| & 9.21 & \textsuperscript{*}\upcell{22.37}{13.16} & 13.07 & \textsuperscript{*}\upcell{24.85}{11.78} \\
\verb|srn| & 21.22 & \textsuperscript{*}\upcell{37.01}{15.79} & 39.08 &\textsuperscript{*} \upcell{51.03}{11.95} \\
\verb|svc| & 18.43 & \textsuperscript{*}\upcell{44.67}{26.24} & 33.39 & \textsuperscript{*}\upcell{59.44}{26.05} \\
\verb|tpi| & 34.90 & \textsuperscript{*}\upcell{49.84}{14.94} & 49.35 & \textsuperscript{*}\upcell{51.66}{2.31} \\
\verb|wes| & 21.24 & \textsuperscript{*}\upcell{51.50}{30.26} & 53.48 & \textsuperscript{*}\upcell{63.66}{10.18} \\
    \midrule
    Avg. & 16.87 & \upcell{35.71}{18.84} & 34.54 & \upcell{48.79}{14.25} \\
    \bottomrule
    \end{tabular}
    \caption{chrF2++ scores with \texttt{Qwen} on the Kreyòl-MT public test set.
    \textsuperscript{*}$\text{p-value} <0.05$.
    }
    \label{tab:creole_mt}
\end{table}

\subsection{Evaluation Results on Kreyòl-MT}
\label{sec:creole_mt_results}

We report chrF2++ scores for \texttt{Qwen} 512-shot translation experiments in Table~\ref{tab:creole_mt}.
We observe significant improvements for nearly all translation directions, with the exceptions of \texttt{eng}$\rightarrow$\texttt{bah} and \texttt{eng}$\rightarrow$\texttt{gul}.

\section{Details of NLLB fine-tuning}
\label{sec:nllb_finetune}
We fine-tune \texttt{nllb-200-distilled-1.3B} using LoRA adapters with rank 16 appied to all linear modules.
We use a batch size of 32 with 8 gradient accumulation steps and a learning rate of $1\times 10^{-4}$.
The learning rate is linearly warmed up over the first 10\% training steps.

For each language not originally supported by NLLB, we add a new special language token.

To upsample lower-resource languages, we use temperature-based sampling. 
The sampling probability for language $i$ is defined as
\begin{align*}
    p_i = \frac{N_i^{1/T}}{\sum_{j}N_j^{1/T}}
\end{align*}
where $N_i$ denotes the number of training sentence pairs for language $i$, $T$ is the temperature, and $p_i$ is the probability of sampling a sentence pair from language $i$ at each training step.
We set $T=5$.

We train for 7,500 steps, corresponding to approximately 5 epochs, and use the final checkpoint for evaluation.

\section{Computation Resources}

All experiments are conducted on a single A100-80GB GPU.
OCR takes from a few minutes to several hours per book, depending on the number of pages and the script.
VLM-based bitext extraction takes approximately 20 minutes per book.
MT experiments take between 30 minutes and 36 hours per translation direction, depending on the number of input sentences and bitext examples.

\end{document}